\documentclass[format=acmsmall, screen=true, review=false]{acmart}
\usepackage[utf8]{inputenc}
\usepackage[T1]{fontenc}

\usepackage{amsmath, amsfonts, bm}

\usepackage{amssymb}

\usepackage{microtype}
\usepackage{nicefrac}

\usepackage{graphicx}
\usepackage{xcolor}
  \definecolor{mycolor}{rgb}{0.85,0.95,0.95}
  \definecolor{headercolor}{RGB}{120,120,120}
  \definecolor{shadecolor}{rgb}{0.9,0.9,0.9}

\usepackage{tikz}
  \usetikzlibrary{arrows.meta, fit, calc}

\usepackage{booktabs, tabularx, array, multirow, makecell}
\usepackage{colortbl}
\usepackage{arydshln}

\usepackage{subcaption}
\usepackage{float}          

\usepackage{enumitem}

\usepackage[ruled,vlined,linesnumbered]{algorithm2e}
  \SetKwInput{KwIn}{Input}
  \SetKwInput{KwOut}{Output}

\usepackage[skins,breakable]{tcolorbox} 
\usepackage{mdframed}                   

\usepackage{caption}
\usepackage{titlesec}

\usepackage{footnote}
  \makesavenoteenv{tabular}
  \makesavenoteenv{table}

\usepackage{hyperref}
\usepackage{cleveref}
  \crefname{algorithm}{Algorithm}{Algorithms}

\usepackage{csquotes}

\newcommand{\dquotes}[1]{``#1''}

\newcommand{\XGBoost}{\texttt{XGB}}

\newenvironment{tealsection}[1]{%
  \section{\textcolor{black}{#1}}\begingroup\color{black}%
}{%
  \endgroup
}

\newcommand{\black}[1]{\textcolor{black}{#1}}

\AtBeginDocument{%
  }

\setcopyright{acmlicensed}
\copyrightyear{2025}
\acmYear{2025}
\acmDOI{XXXXXXX.XXXXXXX}
\acmJournal{TIST}

\begin{document}

\title[XAI4LLM. Let Machine Learning Models and LLMs Collaborate]{XAI4LLM. Let Machine Learning Models and LLMs Collaborate for Enhanced In-Context Learning in Healthcare}

\author{Fatemeh Nazary}
\authornote{Corresponding author}
\email{fatemeh.nazary@poliba.it}
\orcid{0000-0002-6683-9453}
\author{Yashar Deldjoo}
\email{yashar.deldjoo@poliba.it}
\orcid{0000-0002-6767-358X}
\author{Tommaso Di Noia}
\email{tommaso.dinoia@poliba.it}
\orcid{0000-0002-0939-5462}
\author{Eugenio di Sciascio}
\orcid{0000-0002-5484-9945}
\email{eugenio.disciascio@poliba.it}
\affiliation{%
  \institution{Polytechnic University of Bari}
  \city{Bari}
  \country{Italy}
}

\renewcommand{\shortauthors}{Nazary et al.}


\begin{abstract}
\black{Clinical decision support systems require models that are not only highly accurate but also equitable and sensitive to the implications of missed diagnoses. In this study, we introduce a knowledge-guided in-context learning (ICL) framework designed to enable large language models (LLMs) to effectively process structured clinical data. Our approach integrates domain-specific feature groupings, carefully balanced few-shot examples, and task-specific prompting strategies. We systematically evaluate this method across seventy distinct ICL designs by various prompt variations and two different communication styles—natural-language narrative and numeric conversational—and compare its performance to robust classical machine learning (ML) benchmarks on tasks involving heart disease and diabetes prediction.}

\black{Our findings indicate that while traditional ML models maintain superior performance in balanced precision-recall scenarios, LLMs employing narrative prompts with integrated domain knowledge achieve higher recall and significantly reduce gender bias, effectively narrowing fairness disparities by an order of magnitude. Despite the current limitation of increased inference latency, LLMs provide notable advantages, including the capacity for zero-shot deployment and enhanced equity.}

\black{This research offers the first comprehensive analysis of ICL design considerations for applying LLMs to tabular clinical tasks and highlights distillation and multimodal extensions as promising directions for future research.}
\end{abstract}





\maketitle

\section{Introduction}

\black{For decades, clinical prediction problems were dominated by traditional machine-learning (ML) pipelines—carefully engineered \dquotes{narrow-expert} systems that ingest highly‐curated, structured data and produce point-solution predictions for a single medical condition. These ML systems leverage highly structured, curated clinical datasets to deliver precise but narrowly scoped decision support~\cite{stark2019predicting,yuan2024automated,sarkar2020machine}. In contrast, large language models (LLMs), such as ChatGPT~\cite{nazary2023chatgpt,openai2022chatgpt} and Gemini~\cite{team2023gemini,li2023prompt}, have recently demonstrated remarkable versatility through their capacity for generalized reasoning across diverse domains. By utilizing extensive pre-training on vast, heterogeneous datasets and exhibiting emergent behaviors, LLMs introduce the potential for adaptive, context-aware, and flexible clinical decision-making support systems.}

\black{Notwithstanding their great success and potential to inspire new ways of thinking, the path from a general-purpose language model to a trustworthy clinical diagnostic assistant is fraught with significant methodological and ethical challenges. Clinical practice imposes stringent requirements on predictive systems, each presenting distinct obstacles for LLM integration:}

\begin{itemize}
    \item \black{\textbf{Domain precision.} LLMs are predominantly trained on general-domain, non-clinical text sources. Consequently, they lack inherent grounding in clinical thresholds, standardized laboratory cut-offs, validated risk equations, and nuanced causal pathophysiology that underpin precise medical decision-making.}

    \item \black{\textbf{Interpretability.} Clinical decisions cannot rely solely on opaque predictions. Practitioners require clear and meaningful explanations detailing not just the outcomes predicted by a model, but also the clinical rationale and factors influencing these outcomes.}

    \item \black{\textbf{Fairness and risk-sensitivity.} Clinical predictive systems must rigorously minimize false-negative rates, as missed diagnoses can lead to severe patient harm or death. Additionally, demographic biases embedded in model predictions can perpetuate and amplify existing health inequities, necessitating careful scrutiny and proactive mitigation~\cite{deldjoo2023fairnesschatgpt}.}

  \item \black{\textbf{Robust In-context Learning (ICL).} In-context learning (ICL) refers to the ability of very large language models (LLMs) to learn and generalize new tasks solely from the examples or instructions provided within the prompt, without any parameter updates or retraining~\cite{deldjoo2024review,deldjoo2024recommendation}. The capability of an LLM to perform safely and effectively can vary dramatically depending on how input prompts are structured. Seemingly minor changes in question phrasing or context can shift the behavior of a model from expert-level advice to potentially unsafe recommendations, underscoring the need for carefully engineered and rigorously tested prompting strategies.}
\item \black{\textbf{Tabular numerical data.} Clinical ML datasets are typically structured as tabular numerical data, a format that LLMs may inherently struggle to process accurately. In contrast, traditional ML models excel at handling such data efficiently, raising the essential question of how effectively LLMs can integrate and leverage insights derived from numerical, structured clinical datasets~\cite{nazary2023chatgpt}.}
\end{itemize}

\black{Addressing these critical gaps necessitates a structured integration of domain-specific medical expertise—readily accessible from conventional machine learning (ML) models—with the flexible, generative reasoning capacities of LLMs. This paper explicitly investigates the feasibility and effectiveness of such integration. Specifically, we ask: \textit{Can we translate the feature-level insights provided by classical, explainable ML classifiers into a structured form that LLMs can effectively interpret, thereby achieving reliable diagnostic performance even in zero- or few-shot prediction scenarios?} Note that our goal is not to propose a superior LLM-based model over traditional ML models in every sense, necessarily, but rather to consider different evaluation dimensions and practical challenges. We aim to assess whether an LLM-based clinical system can be trusted and effectively structured to augment existing ML capabilities, ensuring safe and reliable clinical decision support.}

\subsection{Contributions}

\black{To address the challenges inherent in translating general-purpose LLMs into reliable clinical diagnostic assistants, we introduce a set of methodological and practical contributions that collectively establish a robust framework for clinical risk prediction tasks:}
\begin{itemize}
\item \black{\textbf{Multi-layer ICL Framework based on XAI (XAI4LLM).} We propose a novel in-context learning strategy that integrates raw patient data, feature-level insights derived from classical ML explainability methods, and explicit task schemas. This framework is motivated by the observation that traditional ML models, trained on localized datasets, behave similarly to \textit{clinicians who provide recommendations based on specific case attributes}; however, in practice, such recommendations could be substituted or enhanced with additional external \textbf{domain knowledge}. Consequently, we explicitly incorporate feature-level importance derived from top-K ML models into the context used by LLMs. Our framework enables general-purpose, chat-based LLMs to reason transparently (and accurately) over traditionally opaque tabular clinical data without requiring task-specific fine-tuning.} \vspace{0.9mm}
\item \black{\textbf{Dual Patient-Profile Encodings and Communication Ablation.}  
We benchmark two complementary ways of \emph{encoding} a patient record and two ways of \emph{delivering} it to the LLM:}

\begin{enumerate}[leftmargin=*]
\item \black{\textbf{Numerical-Conversational (NC).} Every laboratory or clinical measurement is given verbatim as \textit{<feature name>: <exact value>}.}  
      \begin{itemize}
      \item \black{\textbf{NC-ST (single turn).} The entire numeric vector is sent in one message—efficient but cognitively heavy for the LLM.}  
      \item \black{\textbf{NC-MT (multi-turn).} The same vector is split over several quick user–LLM exchanges, allowing the model to reason incrementally (much like a clinician who asks follow-up questions while reviewing vitals).}
      \end{itemize}

\item \black{\textbf{Natural-Language (NL).} The identical record is paraphrased into concise clinical prose (e.g., “A 54-year-old male with non-anginal chest pain, BP 150 mmHg, cholesterol 223 mg/dL…”).}  
      \begin{itemize}
      \item \black{\textbf{NL-ST (single turn).} The full narrative arrives at once—leveraging the LLM’s pre-training on continuous text, but sensitive to prompt length.}
      \end{itemize}
\end{enumerate}

\vspace{0.9mm}

\item \black{\textbf{Reasoning-Mode Control and Few-Shot Curriculum.} 
Our framework rigorously tests two primary reasoning modes—direct inference and CoT—by embedding clear, system-level instructions. Additionally, we introduce a balanced few-shot ICL setup in which LLMs are provided with both positive and negative clinical cases. This approach effectively mitigates class-imbalance biases, even in extremely limited-shot settings (\S\ref{sec:riskaware-objective}).}

\vspace{0.9mm}

\item  \black{\textbf{Explainability-Driven Domain Knowledge and Evaluation Toolbox.}
We leverage hyper-tuned classical models (Logistic Regression, Random Forest, Gradient Boosting, SVM, XGBoost, LightGBM) to derive per-feature importances. These features are systematically bucketed into dominant, moderate, and minor groups using a quantile-based (33\%-67\%-100\%) approach, with only the dominant and moderate categories explicitly incorporated into LLM prompts. Additionally, a comprehensive Python evaluation pipeline automates ML baseline training, bucket extraction, prompt generation (over 70 variants per patient), robust API interactions with automatic retries, and rigorous metric logging with bootstrapped confidence intervals for accuracy, precision, recall, F$_1$, and F$_3$.}

\item \black{\textbf{Risk-Aware Metrics, Fairness Audit, and Key Findings.}
Given the critical nature of clinical prediction tasks, we prioritize the recall-weighted F$_3$ metric to heavily penalize false negatives. We further conduct detailed fairness evaluations, assessing Equality-of-Opportunity, Predictive-Parity, and Predictive-Equality gaps across gender groups. Empirical results demonstrate that an 8-shot numerical single-turn (NL-ST) conversational prompt with Chain-of-Thought (CoT) reasoning achieves an F$_3$ of 0.923 in heart-disease prediction, outperforming the best ML baseline (Gradient Boosting, 0.859), while attaining an F$_1$ within five percentage points of state-of-the-art ML models. Multi-turn conversational prompts (NC-MT) enriched with domain knowledge maximize precision but slightly sacrifice recall, illustrating an inherent precision-recall trade-off. Furthermore, fairness analyses reveal that LLM variants substantially mitigate demographic disparities, achieving near-zero gaps in demographic parity and true positive rates (TPR), unlike traditional ML models that exhibit notable biases disadvantaging female patients. Thus, LLMs not only reduce precision-related biases but significantly enhance gender-specific recall, closely matching ML precision while effectively minimizing false-negative outcomes.}

 \item \black{\textbf{Open-Sourced and Reproducible Framework.} To facilitate transparency and further research, all code, datasets, scripts (for training, explanation extraction, prompt generation, and evaluation), as well as example outputs, are fully open-sourced. This enables the broader community to extend, replicate, or adapt our framework to additional clinical tasks, datasets, and language models.}

\end{itemize}

\black{We explore and validate our approach concretely in the context of heart-disease risk prediction, an archetypal structured clinical decision task. However, the proposed framework and findings generalize naturally and broadly to a wide range of tabular clinical prediction problems.}

\vspace{2mm}

\begin{figure}[!t]
    \centering
    \includegraphics[trim=0.1cm 3.8cm 0.2cm 0.2cm, clip, width=1.0\linewidth]{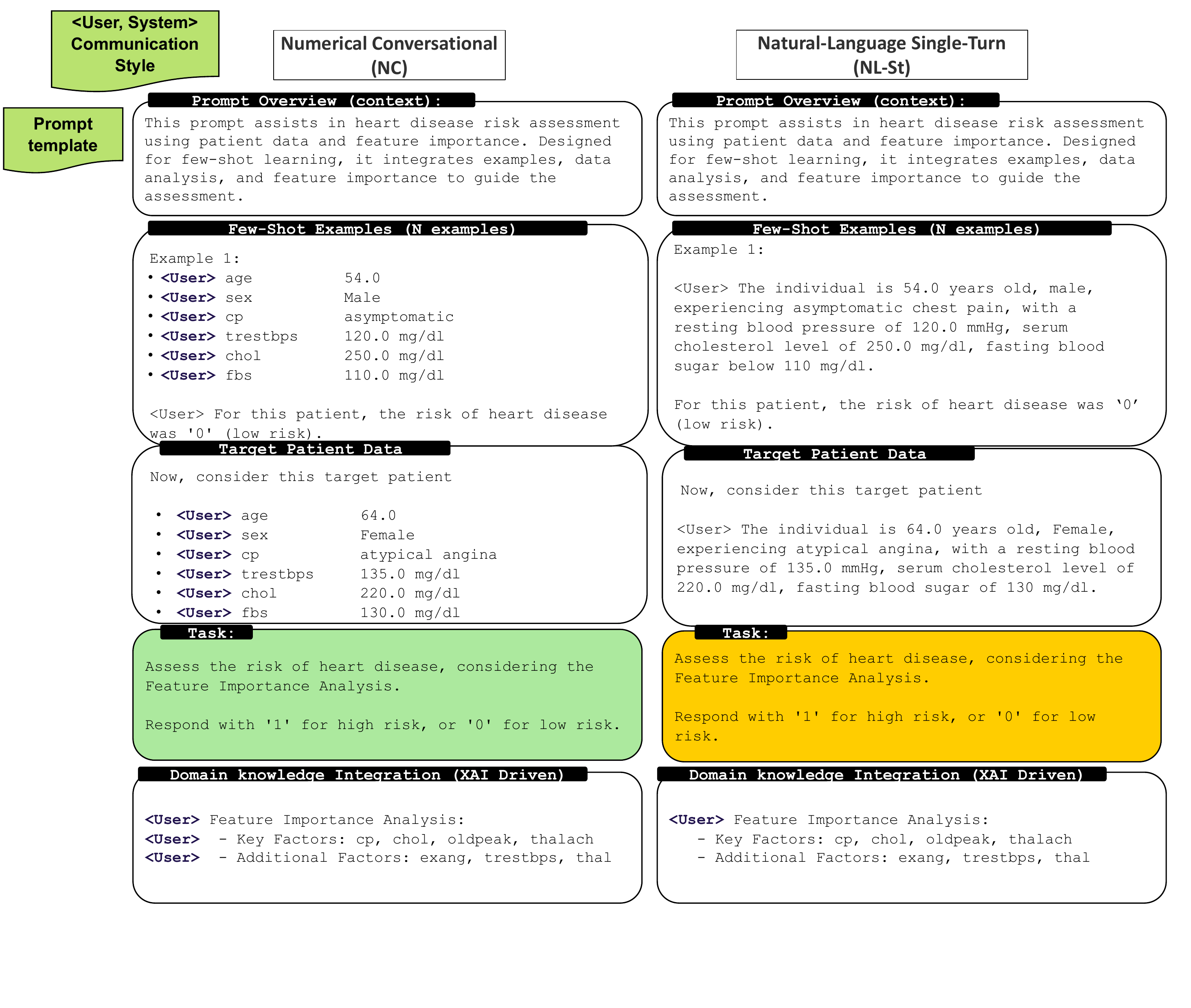}
    \caption{Comparison of two communication syles with LLMs as introduced in this research}
    \label{fig:prompt_style}
\end{figure}

\noindent \textbf{Illustrative Example.} In Figure~\ref{fig:prompt_style}, we illustrate the essence of our contributions through a visual comparison of two distinct communication styles within the context of LLM-based diagnostics. The left column showcases the \dquotes{Conversational} style, where the interaction unfolds through a series of multi-turn exchanges. This method mimics the incremental and dynamic nature of clinician-patient conversations, gradually compiling a comprehensive user profile that blends both numerical data and natural language descriptions. It is designed to engage the LLM in a way that closely parallels the iterative diagnostic process in clinical settings. Conversely, the right column demonstrates the \dquotes{Single-Turn Prompt} approach, which compiles all relevant patient information into one extensive narrative. This style aims to provide the LLM with a complete, integrative overview of the situation of the patient in one round, similar to a clinician reviewing the entire medical history of a patient before making a diagnosis. The contrast between these columns highlights our systematic structured prompt construction ability to adapt the presentation of patient data to optimize LLM-based diagnostic accuracy.

\section{Related work}

\subsection{Generative models and LLMs}

\noindent Generative models have been a cornerstone of machine learning (ML) for some time, showcasing notable success across various applications within the field. Their fundamental concept revolves around understanding the underlying distribution of the data they are trained on—and sampling from it—in order to generate new, similar data. This can range from text and images to more complex outputs. Numerous architectures have been developed, each with its own unique method for learning and reproducing data distributions~\cite{harshvardhan2020comprehensive}. 
These include GANs\footnote{Generative Adversarial Networks}, VAEs\footnote{Variational Autoencoders}, Diffusion Models\footnote{A class of generative models that simulate a diffusion process to gradually transform noise into data}, Conditional Generative models\footnote{Models that generate data conditioned on certain inputs} etc.

Recently, especially after 2023, the popularity of these models has experienced a significant surge, reminiscent of the rise in deep neural networks around 2015 following the introduction of Convolutional Neural Networks (CNNs)~\cite{li2021survey}. This increase in interest is largely due to the advent of large pre-trained language models (LLMs), such as ChatGPT, along with numerous successors such as LLaMA, Claude, Cohere, LaMDA, PaLM, and GPT-Neo, among others \cite{zhao2023survey,xi2023rise}, which have collectively pushed the boundaries of what pre-trained LLMs can achieve in terms of understanding and generating human-like text.

In the context of LLMs, we can recognize two main categories of generative models, each exemplified within healthcare domain.

\begin{enumerate}
    \item \textbf{Directly Trained Models.} These models are explicitly trained on specific datasets, such as patient symptoms, test results, and health outcomes, to learn and sample from the distribution of patient data~\cite{DBLP:conf/flairs/TorfiF20,DBLP:conf/mlhc/ChoiBMDSS17}. This approach benefits from closely aligning the model with the unique characteristics of individual health datasets, enabling predictions of patient outcomes with a detailed comprehension of their medical histories.
    \item \textbf{Pretrained Generative Models.}  These models utilize the vast knowledge acquired from being trained on a broad spectrum of data, textual, visual (and possibly tabular). By fine-tuning these models on disease-specific or demographic-specific datasets, their capacity to accurately interpret complex medical scenarios and patient histories is enhanced, thus improving their ability to offer precise diagnostic insights or personalized treatment plans. They can be used in the following scenarios~\cite{DBLP:journals/jbi/GeGDAS23,DBLP:conf/ml4h/MoorHWYDLZRR23,DBLP:journals/mta/SpolaorLMNPTCWF24,DBLP:journals/artmed/HenrikssonPHN23,shentu2024cxr,he2023large}:

    \begin{itemize}
\item[a)] \textbf{Zero- and Few-shot In-Context Learning.} This represents a remarkable emergent capability of LLMs, showcasing exceptional zero- and few-shot ICL. For example, these techniques enable machine learning models to apply previously learned concepts to novel problems without the need for extensive further training. This is particularly important in emerging health issues, where data may be limited; 
\item[b)] \textbf{Fine-Tuning.} This process involves adjusting pre-trained models with in-domain data, such as targeted medical datasets, thereby improving their effectiveness for specific health conditions and enhancing the personalization of care.
\item[c)]\textbf{ Embedding Generation for Downstream Training.} This technique is particularly useful for interpreting
complex medical data such as electronic health records or imaging data, where embeddings can capture intricate
patterns and relationships \cite{yang2023large,li2023prompt}.
\end{itemize}
\end{enumerate}

As the role of AI/ML in critical domains such as healthcare continues to mature, it is important to understand the challenges of integrating these new technologies into clinical practice, and understand their associated risks. Our work at hand is focused on the first category here, zero- and few-shot ICL.

\subsection{Overview of LLMs in the healthcare}
\label{subsec:rw_health}
Large Language Models (LLMs) and, more broadly, generative AI, hold the potential for widespread application across various domands and task, including finance~\cite{wu2023bloomberggpt,deldjoo2023fairnesschatgpt}, healthcare~\cite{nazary2023chatgpt,peng2023study,ghosh2024clipsyntel}, agriculture~\cite{rezayi2023exploring}, recommender systems~\cite{deldjoo2024fairevalllm,deldjoo2024review} and more, as explored in recent research works. Of particular importance are critical sectors such as healthcare and finance, where the societal impact is significant. In these areas, it is crucial not only to focus on the positives and benefits offered by the system in terms of better predictions but also to critically understand and evaluate the associated risks and challenges.

The healthcare sector is one of the most critical areas where LLMs or more broadly generative AI could have a transformative effect, but it also poses  risks since decisions directly impact human lives. On the positive side, a number of recent surveys~\cite{yang2023large,nazi2023large,chen2024generative,umer2024generative} provide a good frame of refrence on various tasks that LLMs can achieve within the healthcare sector. These include medical assessment, medical question answering, scientific writing, eHealth care, and patient classification.

For instance, the survey by~\citet{yang2023large} examines the potential applications of LLMs in healthcare, highlighting their role in developing domain-specific models to enhance clinical utility. These models aim to minimize errors and maximize effectiveness for global healthcare delivery, either by building from scratch or refining existing larger models. This approach facilitates decentralized research, enabling researchers worldwide to tailor LLM applications to specific clinical tasks. Moreover, the authors detail how LLMs integrate into the healthcare process, which is categorized into three main stages: (1) \textbf{Pre-consultation}, involving screening and initial medical consultations for symptomatic or asymptomatic patients; (2) \textbf{Diagnosis}, which includes collecting patient history, conducting physical examinations, and performing diagnostic investigations such as CT scans; and (3) \textbf{Management}, encompassing the prescription of medications, patient education, counseling, and management of insurance claims. The approach of examining LLM applications according to \textit{timelines} and \textit{temporal screening} presents an interesting method for categorizing these interventions.

Another notable survey by \citet{nazi2023large} presents an alternative interesting categorization of the various applications of LLMs in healthcare, organized for different stakeholders: for (1) \textbf{patients}, LLMs support health education, lifestyle advice, patient care, and function through virtual medical assistants and health chatbots; for (2) \textbf{doctors}, they extend to treatment planning, collaborative diagnosis, medical literature analysis, and clinical decision support; and for (3) \textbf{central healthcare applications}, they include drug discovery, automated medical report synthesis, and radiology and imaging. This structured approach underscores the wide-ranging impact of LLMs in enhancing and personalizing healthcare services across different levels of medical practice.

\subsection{LLMs for Automated Healthcare Diagnosis}

The integration of LLMs into healthcare has been an area of significant research interest, aiming to enhance diagnostic accuracy and decision-making processes \cite{sayin2024can,kusa2023dr, galitsky2024llm, goyal2024healai}. The work at hand contributes to advancing the application of LLMs in medical diagnostics, each adopting unique approaches to harness the capabilities of LLMs in addressing complex healthcare challenges. A few recent research works are reviewed in the following:

The research work by \citet{sayin2024can} emphasize the role of LLMs in supporting medical decision-making but focuses on correcting physician errors during diagnostic processes. This research highlights the importance of prompt design in the performance of LLMs, similar to our approach. However it has a specific focus on interactive scenarios between physicians and LLMs, aimed at improving diagnostic outcomes based on physician-LLM collaboration. \citet{kusa2023dr} explore the impact of user beliefs and prompt formulation on health diagnoses, demonstrating the sensitivity of LLMs to input variations. This study complements our findings by addressing the challenges in ensuring consistency in LLM-generated diagnoses, emphasizing one the  role of tailored prompt design to mitigate the impact of user input biases.
\citet{galitsky2024llm} studies the personalization in LLM applications, particularly in generating tailored medical recommendations. While our focus is on diagnostic accuracy, the work by Gaitsky et al. underscores the potential of LLMs in personalizing healthcare interactions, which could enhance the user-specific applicability of diagnostic tools. \citet{gramopadhye2024few} explore chain-of-thought (CoT) reasoning in LLMs for medical question answering, a technique that is also used in our work together with the structured prompt design by facilitating detailed, step-by-step reasoning processes in LLMs. This approach aims at improving the interpretability and accuracy of LLM responses, which aligns with our goal to enhance diagnostic precision through detailed data analysis and reasoning. \vspace{2mm}

\black{\noindent \textbf{Positioning \emph{XAI4LLM} within the Literature.} Early explorations of LLMs in clinical settings have followed two main paths. First, some works treat an LLM as a stand-alone diagnostic agent, evaluating it \dquotes{out-of-the-box} on medical exam-style benchmark vignettes and board–style examinations~\cite{sayin2024llmscorrectphysiciansyet,nori2023capabilities,griot2025large}. These studies showcase the impressive emergent reasoning of foundation models – for example, GPT-4 exceeded the USMLE passing score by over 20 points without specialized training~\cite{nori2023capabilities}, and models such as Med-PaLM reached expert-level accuracy on multiple-choice medical questions~\cite{singhal2023large}. However, they also highlight substantial limitations. When tasks require parsing highly structured, numeric data or dealing with class imbalance (e.g., coding diagnoses from tabular records), LLM performance drops off markedly~\cite{griot2025large}. In such cases, classical machine-learning (ML) models (e.g., gradient boosting on tabular features) remain state-of-the-art. In other words, vanilla LLMs excel at open-ended text reasoning but often struggle with structured clinical data that traditional pipelines handle well.}

\black{On the other hand, a newer research stream integrates LLMs within a broader clinical decision-support stack, rather than using them alone. Typical approaches include: (i) fine-tuning domain-specific LMs (e.g., SciBERT or other bioclinical BERT/GPT variants) on medical texts such as discharge summaries
\cite{singhal2023large} to leverage biomedical corpora; (ii) prompt-engineering techniques to elicit chain-of-thought rationales for clinical reasoning \cite{kojima2022large,savage2024diagnostic} (for example, instructing the LLM \dquotes{Let’s think step by step} to guide differential diagnosis reasoning); and (iii) retrieval-augmented generation (RAG) systems that feed the LLM with relevant clinical guidelines or literature excerpts alongside its free-text advice \cite{amugongo2025retrieval}. These hybrid strategies have been shown to boost factual accuracy and traceability – e.g., grounding the answers of an LLM in up-to-date clinical guidelines yields more personalized and correct recommendations. Nevertheless, in most such setups, the LLM is still treated as a black-box oracle: its final predictions are accepted or rejected in full. There is usually no explicit fusion of the reasoning abilities of LLMs with the feature-level knowledge from pre-existing ML risk models. In other words, the human or system must choose between the LLM’s answer and the classical model, rather than combining their insights.}

\black{The differences XAI4LLM makes include:}
\begin{itemize}
    \item  \black{\textbf{Hybrid, not replacement.}  
Unlike prior work that implicitly compares LLMs against conventional models, XAI4LLM treats them as collaborators.\textit{ Our goal is not to show a generic LLM outperforming a purpose-built tabular model, but to transfer the clinically validated signal from the tabular model into the prompt constructed for an LLM}, so that both components work in harmony. To our knowledge, this is the first empirical study to systematically quantify how incorporating quantile-bucketed feature importances (exported from an ensemble of interpretable classifiers) affects the downstream task performance (measured by accuracy, risk, and fairness-oriented metrics).}

\item \black{\textbf{Communication as a Design Variable.} Prior LLM-in-health studies overwhelmingly rely on single-shot narrative prompts. In contrast, we deliberately vary the dialogue format based on cognitive load principles~\cite{sweller2019cognitive}. We compare a numerical multi-turn (conversational) prompt – feeding patient data in small, structured chunks over several turns – against a traditional natural-language single-turn summary of the case. Our ablation finds that the multi-turn numeric format attains ML-level accuracy under tight token budgets, whereas the single-turn narrative shines in low-shot scenarios. This trade-off (efficiency vs. expressiveness) was not examined in earlier work, which typically fixed the prompt style. It highlights how we communicate information to an LLM can be as important as what we communicate.}

\item \black{\textbf{Risk-aware Evaluation.} Most benchmarks for clinical AI stop at overall accuracy or AUROC, which can obscure models’ behavior on rare but critical cases. We contribute an open-source evaluation toolkit that reports recall-weighted $F_{3}$ (placing heavy weight on recall to penalize missed diagnoses), false-negative rates, and several fairness gap metrics (e.g. demographic parity, TPR parity, FPR parity). This emphasis on cost-sensitive and equity-oriented metrics aligns with recent calls for responsible AI in medicine. By examining metrics like gaps in true positive rate between male vs. female patients, we go beyond contemporaneous prompt-engineering papers that often overlook demographic performance. Our results show, for instance, that the LLM integration can substantially reduce gender bias in recall compared to a baseline ML model, addressing the kind of issues raised by \cite{obermeyer2019dissecting} regarding algorithmic bias in healthcare.}

\item \black{\textbf{Risk-aware evaluation.}  
Most existing benchmarks stop at accuracy or AUROC.  
We contribute a reproducible python toolbox that reports recall-weighted $F_{3}$, false-negative rate, and three fairness gaps (DP, TPR, FPR).  
This emphasis on \emph{cost-sensitive} and \emph{equity-oriented} metrics aligns with recent calls for responsible AI in medicine \cite{rajkomar2018ensuring, obermeyer2019dissecting} and differentiates our study from contemporaneous prompt-engineering papers that overlook demographic performance.}

\item \black{\textbf{End-to-end openness.}  
Whereas many prior prototypes rely on proprietary code or private data, we release the entire pipeline—from hyper-parameter search to prompt generation and evaluation—under an open-source licence, facilitating direct comparison and extension to new tabular tasks such as chronic-kidney failure or sepsis triage.}

\end{itemize}

\black{In combination, these four facets place \emph{XAI4LLM} at the intersection of explainable ML and generative reasoning: we treat LLMs \textit{as cooperative partners} that can be steered, audited, and ultimately trusted through carefully structured conversations seeded with explicit domain knowledge.  The positive empirical results on heart-disease prediction suggest that this paradigm is a viable blueprint for other high-stakes, tabular-heavy areas of clinical AI.}

\section{Foundations}
\noindent This section will briefly discuss the foundations of this work, contrasting generative versus discriminative models (Section~\ref{subsec:gen_vs_desc}), and connecting this to the approach of LLMs (Section~\ref{subsec:LLM}). It would basically help us understand the application of LLMs using prompt-based ICL for binary classification.

\subsection{Generative and Discriminative Models}
\label{subsec:gen_vs_desc}
\noindent Generative and Discriminative models are two main methods in pattern recognition and machine learning tasks. Their fundamental difference lies in their approach to modeling and application:

\begin{itemize}
    \item \textbf{Discriminative Models:} These models focus on the probability of a label ($y$) given an observable (features $x$), formalized as:
    \begin{equation}
        P(Y|X = x)
    \end{equation}
    These models are widely used for prediction tasks, such as classification or regression. Discriminative models estimate the likelihood of a label based on observed features, effectively discriminating between different kinds of data instances.
    \item \textbf{Generative Models:} These models aim to model the distribution of observables within a dataset $\mathcal{X}$ given a label $y$. This is formally represented as:
    \begin{equation}
        P(X|Y = y)
    \end{equation}
    Such models are thus proper for generating new data instances by learning the joint probability distribution $P(X, Y)$ or simply $P(X)$ when labels are not available. For instance, after learning the general characteristics of cats from a dataset, a generative model can create a new image that captures the essence of a cat, as it understands the joint probability distribution of the data and labels, $P(X, Y)$. This capability can be expressed as:

\begin{equation}
X_{\text{new}} \sim P(X|Y = \text{"cat"})
\end{equation}

Here, $X_{\text{new}}$ represents a new instance (e.g., an image of a cat) generated by the model, drawing from its learned distribution of features associated with cats. Thus, generative models can not only understand the characteristics that define each class but also produce new examples that reflect those insights.

\end{itemize}

\subsection{Utilizing pLLMs in Discriminative Tasks via ICL}
\label{subsec:LLM}
This section aims to bridge the conceptual gap between the probabilistic approaches of discriminative and generative models and their application within LLMs, especially in the context of binary classification tasks.

Given a dataset $\{(x_i, y_i)\}_{i=1}^N$ where $x_i \in \mathcal{X}$ represents the input features and $y_i \in \{0, 1\}$ denotes the binary labels, the primary goal of a learning model is to accurately predict the label of unseen data points through a learned mapping $f$. This process typically involves minimizing a loss function, such as cross-entropy loss, to reduce the discrepancy between the predicted and actual labels.

\subsubsection{Discriminative approach to binary classification}
Discriminative models, or direct learning models, seek to map input features $X$ to labels $Y$ via a function $f: X \rightarrow Y$.
In this approach, the objective of the models --often referred to as a \textit{learning model}-- is to learn the mapping $f$ to accurately predict the label of unseen data points. This is achieved by minimizing a loss function, such as cross-entropy, which quantifies the discrepancy between the predicted labels and the actual labels.

\subsubsection{Generative pLLMs' approach to binary classification}
In contrast, generative models, herein referred to as \dquotes{\textbf{foundation models,}} employ a distinct tactic. Instead of directly mapping inputs to labels, they aim to comprehend and replicate the data distribution to generate new, plausible data points. For binary classification, these models are adapted through prompt-based techniques to function as classifiers:
\begin{equation}
    G: (prompt(x), \theta) \rightarrow y
\end{equation}
Here, $prompt(x)$ is a mechanism that reformats the input $x$ into a configuration amenable to the foundation model $G$, integrating the binary classification task within the operational framework of the model. Parameterized by $\theta$, the model subsequently produces a response construed as the class label $y$. Thus, this approach leverages the generative capabilities of the model to perform classification, relying on the internal understanding of the model from the task as framed by the prompt.

\begin{tealsection}{Proposed \texttt{XAI4LLM} Framework}
\label{sec:xai4llm}

\subsection{Design Rationale and Motivation}

Modern Large Language Models (LLMs) offer powerful reasoning capabilities derived from massive-scale pretraining on general textual corpora. However, they inherently lack structured clinical reasoning tailored explicitly to medical domains. This gap—between implicit general reasoning and explicit clinical knowledge—is precisely the challenge our \texttt{XAI4LLM} framework addresses. Specifically, we propose augmenting LLM reasoning with structured domain insights distilled directly from interpretable classical machine learning (ML) models. 

We anchor our approach on two complementary principles:

\begin{enumerate}[leftmargin=*]
\item \textbf{Knowledge Alignment:}  
Implicit reasoning of LLMs must be systematically enriched by explicit, structured domain insights (feature importance, clinical heuristics) derived from explainable ML methods.

\item \textbf{Prompt-as-Program Paradigm:}  
We conceptualize prompts as structured programs, whose syntax and communication style with LLM (single-turn vs. multi-turn dialogue) and semantics (numerical vs. narrative textual encoding) can be systematically varied and optimized.  
\end{enumerate}
\vspace{5mm}

\noindent These principles yield a clear two-layer architecture:

\begin{description}[leftmargin=*]
\item[Layer 1 – \textit{Prompt Construction:}]  
Assembles a structured, multi-layer prompt from patient data, few-shot exemplars, and feature-importance-based domain knowledge.

\item[Layer~2 --- \emph{LLM Interaction and Prediction.}] Utilizes the constructed prompts to interact with an off-the-shelf, pretrained LLM, explicitly steering the reasoning process of the model (either via direct inference or chain-of-thought reasoning), and converting its textual response into actionable clinical predictions via robust NLP-based post-processing. \\

We now study these steps in the design choices in more detail, discussing the core involved components:
\end{description}

\subsection{Formalizing the Prompt Construction}

\noindent \textbf{Notation.} Let $\mathcal{D} = \{(x_i, y_i)\}_{i=1}^N$ denote a clinical dataset, where $x_i \in \mathcal{X} \subseteq \mathbb{R}^p$ is the feature vector and $y_i \in \{0,1\}$ represents the binary clinical outcome (the generalization to multi-class outcomes is straightforward). We further denote by $E = \{(x_j, y_j)\}_{j=1}^n$ the set of $n$ few-shot examples, and by $x^\star$ the target patient instance requiring prediction. Classical ML models $\mathcal{M} = \{f_m\}_{m=1}^M$ yield both predictive outcomes $f_m(x)$ and corresponding feature importance vectors $\varphi_m \in \mathbb{R}_{\geq 0}^{p}$. These vectors are distilled into a structured domain-knowledge tuple:
\begin{equation}
K_{\text{XAI}} = \left( \mathcal{F}_{\text{dom}},\, \mathcal{F}_{\text{mod}},\, \mathcal{F}_{\text{min}} \right),
\end{equation}
via quantile-based binning described in Equation~\eqref{eq:bucket}. In particular, given importances $\varphi\in\mathbb{R}_{\ge 0}^p$, tiers are obtained via
\begin{equation}
\label{eq:bucket}
\mathcal{F}_{c}
    =\Bigl\{\,f\in\![p] \;\bigl|\;
        q_{c-1} < \varphi_f \le q_{c}\Bigr\}, \qquad
    c\in\{\text{min},\text{mod},\text{dom}\},
\end{equation}
where $(q_{\text{min}},q_{\text{mod}},q_{\text{dom}})$ are the empirical $(33\%,67\%,100\%)$ quantiles.

\paragraph{Prompt Generator.}
We define a deterministic prompt-generation function:
\begin{equation}
\label{eq:prompt}
\mathcal{P}\!:\! \left(x^\star,\, E,\, K_{\text{XAI}},\, \theta_{\text{comm}},\, \theta_{\text{reason}}\right)
    \longmapsto \textsf{Prompt} \in \mathbb{T}^{\leq L},
\end{equation}
where $\mathbb{T}^{\leq L}$ represents the space of allowable textual tokens (with length constraint $L$), and hyperparameters $\theta_{\text{comm}} \in \{\textsc{NC-MT},\, \textsc{NL-ST}\}$ and $\theta_{\text{reason}} \in \{\textsc{Direct},\, \textsc{CoT}\}$ specify the communication style and reasoning mode, respectively. Internally, the prompt is structurally composed of five sequential blocks:
\begin{align}
\mathcal{P}(\cdot) =\, &
    \underbrace{\textsf{Intro}(T)}_{\text{Task preamble}}
    + \underbrace{\textsf{Domain}(K_{\text{XAI}})}_{\text{Clinician-guided domain knowledge}}
    + \underbrace{\textsf{Shots}(E)}_{\text{Few-shot examples}}
    \nonumber\\
    & + \underbrace{\textsf{Profile}(x^\star)}_{\text{Patient data representation}}
    + \underbrace{\textsf{TaskInst}(I)}_{\text{Structured JSON schema instruction}}.
\end{align}

\paragraph{End-to-End Inference.}
Given a pretrained language model operator $\mathcal{C}_{\text{LLM}}$ (e.g., GPT-3.5-Turbo), the final clinical risk prediction is computed as:
\begin{equation}
\hat{y}(x^\star) =
    \psi \left( \mathcal{C}_{\text{LLM}} \left[\, \mathcal{P}(x^\star,\, E,\, K_{\text{XAI}},\, \theta_{\text{comm}},\, \theta_{\text{reason}})\, \right] \right),
\end{equation}
where $\psi : \mathbb{T}^{\leq 2048} \rightarrow \{0,1\}$ robustly parses the LLM output into a binary clinical decision.

\subsection{Communication Styles (\texorpdfstring{$\theta_{\text{comm}}$}{}) with LLM.}
\label{subsec:com_style}

The encoding of patient profiles dramatically influences how effectively an LLM can process and interpret clinical information. We systematically explore three communication styles, each with distinct cognitive and computational implications:

\begin{enumerate}[leftmargin=*]
\item \textbf{Numerical Conversational Single-Turn (NC):}  
A concise, single-step enumeration of numerical feature values, preserving the fidelity of exact measurements.

\textit{Example:}
\begin{verbatim}
Age: 54, Sex: 1, CP: 3, BP: 150 mmHg, Chol: 223 mg/dL, ...
\end{verbatim}

\item \textbf{Numerical Conversational Multi-Turn (NC-MT):}  
Features are introduced incrementally in a conversational exchange, emulating clinical triage. Each data point prompts a minimal acknowledgment from the LLM, providing room for intermediate reasoning steps—crucial in CoT reasoning.

\textit{Example Dialogue:}
\begin{verbatim}
<User> Age: 54
<LLM> Noted.
<User> Sex: 1
<LLM> Noted.
...
\end{verbatim}

This style reduces the token-overload inherent in single-turn numerical presentations, improving the coherence and interpretability of LLM reasoning.

\item \textbf{Natural-Language Single-Turn (NL-ST):}  
A rich, clinical-style narrative synthesizing multiple patient features, benefiting from LLM training on large amounts of coherent text: \\

\textit{Example:}
\begin{quote}
\textit{A 54-year-old male presenting with non-anginal chest pain, resting blood pressure of 150 mmHg, and serum cholesterol of 223 mg/dL…}
\end{quote} \vspace{2mm}

Please note that NL-ST leverages narrative understanding but risks semantic compression and the potential loss of numerical precision.
\end{enumerate} \vspace{3mm}

These encodings explicitly test the trade-off between precision, readability, and computational feasibility, enabling systematic optimization of the prompt-based interaction. \textit{Please note that in the real experiment, due to time constraints and based on sample testing, we report all results using NC-MT and NL-ST, as they provided better quality.}

\subsection{Reasoning Modes and the Chain-of-Thought Paradigm (\texorpdfstring{$\theta_{\text{reason}}$}{})}

We introduce two explicit reasoning modes to guide the cognitive processes of the LLM:

\begin{enumerate}[leftmargin=*]
\item \textbf{Direct reasoning:} The LLM directly infers the binary outcome from the provided context without revealing intermediate steps. This mode tests the pure predictive ability of implicit LLM reasoning.

\item \textbf{Chain-of-Thought (CoT) reasoning:} The LLM explicitly verbalizes intermediate reasoning steps before reaching a final binary decision. A structured delimiter (\texttt{ANSWER\_JSON:}) ensures the clear extraction of the binary prediction.  

\textit{Example of CoT output:}
\begin{quote}
\textit{"Given the patient's chest pain type, elevated cholesterol, and high blood pressure, the clinical picture suggests elevated risk. Thus, my final prediction is as follows:\\
ANSWER\_JSON: \{“risk”: 1\}"}
\end{quote}
\end{enumerate}

Chain-of-Thought reasoning facilitates model transparency, clinical validation, and the ability to audit decision-making pathways within LLM predictions.

\subsection{Clinical Risk-Aware Objective (Revised)}
\label{sec:riskaware-objective}

To align evaluation with clinical priorities, we report (i) \textbf{standard, precision-balanced accuracy} (F-score with $\beta=1$), (ii) \textbf{recall-weighted, risk-sensitive accuracy} (F-score with $\beta=3$), and (iii) \textbf{fairness gaps} that quantify demographic disparities in true-positive, positive-predictive, and overall accuracy rates.

\vspace{0.5em}
\noindent
\textbf{Notation.}
\begin{itemize}
    \item TP = true positives \hspace{2em} FP = false positives
    \item FN = false negatives \hspace{2em} TN = true negatives
    \item Precision $= \frac{\text{TP}}{\text{TP} + \text{FP}}$
    \item Recall $= \frac{\text{TP}}{\text{TP} + \text{FN}}$
\end{itemize}

\paragraph{A. F-score Family}
The general F-score with parameter $\beta$ weights recall $\beta^2$ times more than precision:
\begin{equation}
F_\beta = \frac{(1+\beta^2) \cdot \text{Precision} \cdot \text{Recall}}{\beta^2 \cdot \text{Precision} + \text{Recall}}
\end{equation}

\textbf{F1 (Balanced)}\,\,($\beta=1$):
\begin{equation}
F_1 = \frac{2 \cdot \text{TP}}{2\,\text{TP} + \text{FP} + \text{FN}}
\end{equation}

\textbf{F3 (Risk-aware)}\,\,($\beta=3$):
\begin{equation}
F_3 = \frac{10 \cdot \text{Precision} \cdot \text{Recall}}{9\,\text{Precision} + \text{Recall}}
= \frac{10 \cdot \text{TP}}{10\,\text{TP} + 9\,\text{FP} + \text{FN}}
\end{equation}

F3 penalizes false negatives \textbf{three times} more than false positives, reflecting the higher clinical cost of missed diagnoses.

\paragraph{B. Fairness Gaps}
Let the sensitive attribute take two values (e.g., $M$ = male, $F$ = female). For each group $g$ we compute:
\begin{align*}
\text{True-positive rate (TPR):}\quad & \mathrm{TPR}_g = \frac{\text{TP}_g}{\text{TP}_g + \text{FN}_g} \\
\text{Positive-predictive value (PPV):}\quad & \mathrm{PPV}_g = \frac{\text{TP}_g}{\text{TP}_g + \text{FP}_g} \\
\text{Accuracy (ACC):}\quad & \mathrm{ACC}_g = \frac{\text{TP}_g + \text{TN}_g}{N_g}
\end{align*}

\noindent
The following absolute-gap metrics summarize cross-group disparities ($0=$ perfectly fair):
\begin{align}
\Delta\text{EO}      &= \left| \mathrm{TPR}_\mathrm{M} - \mathrm{TPR}_\mathrm{F} \right| \quad &&\text{(Equality of Opportunity)} \\[0.5em]
\Delta\text{PP}      &= \left| \mathrm{PPV}_\mathrm{M} - \mathrm{PPV}_\mathrm{F} \right| \quad &&\text{(Predictive Parity)} \\[0.5em]
\Delta\text{PEAcc}   &= \left| \mathrm{ACC}_\mathrm{M} - \mathrm{ACC}_\mathrm{F} \right| \quad &&\text{(Predictive Equality – Accuracy)}
\end{align}

A model satisfying $\Delta\text{EO} \approx 0$ ensures that males and females experience comparable sensitivity (recall) and therefore similar protection against missed diagnoses. The additional parity measures provide complementary lenses on precision- and accuracy-based fairness.

\paragraph{Implementation note.} All counts are computed on the \textbf{held-out test set}; confidence intervals are obtained via 1,000-fold non-parametric bootstrap.

\vspace{0.5em}

Algorithm~\ref{alg:xai4llm}  provides explicit pseudocode that implements the described pipeline end-to-end. \end{tealsection}

\begin{algorithm}[H]
\caption{\textsc{XAI4LLM-Inference}}\label{alg:xai4llm}
\small
\KwIn{Target patient $x^\star$; few-shot set $E$; ML ensemble $\mathcal{M}$;\\
\hspace{1.6em} hyper-parameters $(\theta_{\text{comm}},\theta_{\text{reason}})$}
\KwOut{Risk flag $\hat{y}\in\{0,1\}$}

\vspace{0.2em}
\tcp{1. Extract structured domain knowledge}
Compute feature importances $\{\varphi_m\}_{m\in\mathcal{M}}$\;
$K_{\!\text{XAI}}\gets$ \textsc{QuantileBucket}$(\{\varphi_m\})$\;

\vspace{0.2em}
\tcp{2. Assemble prompt}
$P\gets\mathcal{P}(x^\star,E,K_{\!\text{XAI}},\theta_{\text{comm}},\theta_{\text{reason}})$\;

\vspace{0.2em}
\tcp{3. Query frozen LLM}
$r\gets\mathcal{C}_{\text{LLM}}[P]$\;

\vspace{0.2em}
\tcp{4. Parse reply \& return prediction}
$\hat{y}\gets\psi(r)$\;
\Return $\hat{y}$\;
\end{algorithm}

\begin{figure}
    \centering
    \includegraphics[width=0.65\linewidth]{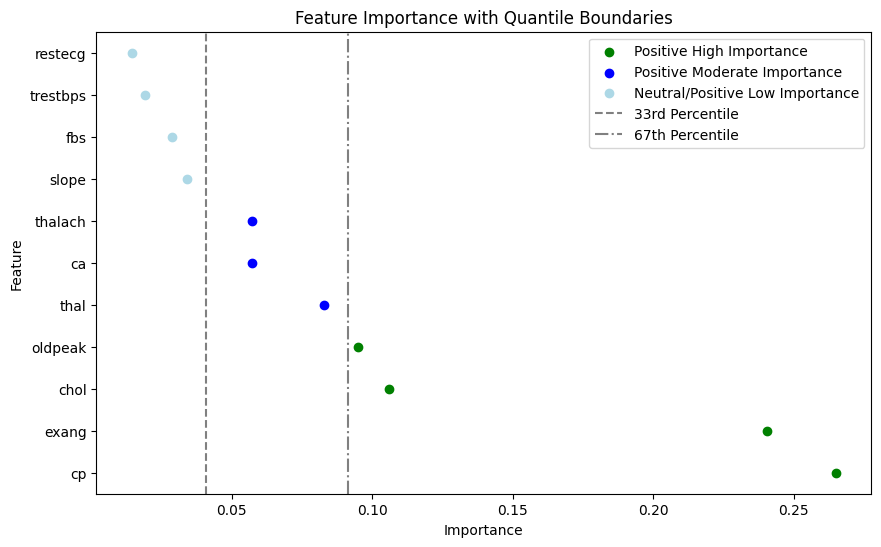}
    \caption{Scatter plot of feature importance with quantile boundaries, depicting the varying influence of \XGBoost method on heart disease risk assessment. The color coding—green for high importance, blue for moderate, and light blue for low importance—alongside percentile markers, provides a clear hierarchical visualization of the factors contributing to the prediction of the model.}
    \label{fig:FI}
\end{figure}

\section{Experimental Setup}

This study examines binary clinical-risk prediction across three main factors: dataset imbalance (in both sex and outcome prevalence), communication style (numeric multi-turn (\texttt{NC-MT}) vs. natural-language single-turn (\texttt{NL-ST})), and reasoning mode (direct answers vs. chain-of-thought (\texttt{CoT})). Formally, we learn a mapping $f: \mathcal{X} \rightarrow \{0,1\}$, where $x \in \mathcal{X} \subseteq \mathbb{R}^p$ and $y \in \{0,1\}$.

\vspace{1em}

We use two publicly available clinical datasets, each pre-processed using the same pipeline: rows with 40\% or more missing values are dropped, remaining numeric \texttt{NaN} values are median-imputed, and categorical features with low cardinality are label-encoded. The target variable is binarized where needed (for heart disease), and a stratified 10\% of patients are held out for all downstream experiments. Class and sex distributions are visualized for both datasets.

\begin{table}[h]
\footnotesize
\caption{Key characteristics of the clinical cohorts.}
\centering
\begin{tabular}{lcccc}
    \toprule
    Dataset & Records ($n$) & Attributes ($p$) & Sex ratio & Raw prevalence \\
    \midrule
    \textbf{Heart Disease}  & 920   & 13 & 78\% male / 22\% female & 63\% male / 28\% female \\
    \black{\textbf{Stroke-Prediction}} & \black{5,110} & \black{11} & \black{41\% male / 59\% female} & \black{4.9\% overall} \\
    \bottomrule
\end{tabular}
\label{tab:datasets}
\end{table}

\vspace{1em}

For benchmarking, we report results for two feature-agnostic baselines---a random predictor (drawing from the empirical class prior) and a stratified classifier (\texttt{DummyClassifier} in scikit-learn). Five classical ML models are also evaluated: Logistic Regression, Random Forest, Gradient Boosting, Support Vector Machine, and XGBoost (if installed). Each model is tuned via randomized hyper-parameter search, as detailed in Table~\ref{tab:ml_hyperparams}. Wall-clock times for both training and inference are recorded.

\begin{table}[h]
\footnotesize
\centering
\caption{Hyper-parameter search grids for classical models.}
\begin{tabular}{l p{10cm}}
    \toprule
    Algorithm & Search space (Randomized, 3-fold stratified CV, $\leq$ 20 draws) \\
    \midrule
    Logistic Regression & $C \in \{0.1, 1, 10\}$, penalty $\in$ \{\texttt{l1}, \texttt{l2}\} \\
    Random Forest & $n_{\mathrm{estimators}} \in \{100, 200, 400\}$, max\_depth $\in$ \{\texttt{None}, 5, 10, 20\}, min\_samples\_split $\in$ \{2, 4\} \\
    Gradient Boosting & $n_{\mathrm{estimators}} \in \{100, 200\}$, learning\_rate $\in$ \{0.03, 0.05, 0.1\}, max\_depth $\in$ \{3, 5\} \\
    Support Vector Machine & $C \in \{0.1, 1, 10\}$, kernel $\in$ \{\texttt{rbf}, \texttt{linear}\}, gamma $\in$ \{\texttt{scale}, \texttt{auto}\} \\
    XGBoost  & $n_{\mathrm{estimators}} \in \{50, 100, 200\}$, max\_depth $\in$ \{3, 5, 8\}, learning\_rate $\in$ \{0.03, 0.1, 0.2\}, subsample $\in$ \{0.8, 1.0\} \\
    \bottomrule
\end{tabular}
\label{tab:ml_hyperparams}
\end{table}

\vspace{1em}

We further evaluate OpenAI's \texttt{gpt-3.5-turbo} (June 2025 weights) in a deterministic setting (\texttt{temperature}=0), using batch calls with exponential back-off for error handling. The LLM prompt grid covers combinations of few-shot exemplars ($k \in \{0, 16\}$), communication style (\texttt{NC-MT} vs. \texttt{NL-ST}), reasoning mode (direct or \texttt{CoT}), and the inclusion of domain-knowledge blocks with or without tiered feature hints---for a total of 16 prompt variants per test patient. All prompt templates and parsing code are open-source.

\subsection{Experimental Research Questions}
We designed and conducted a set of experiments to address the following experimental research questions (RQs):
\label{subsec:RQs}

\begin{itemize}

\item \textbf{RQ-1}: How do LLMs compare with classical ML baselines on precision-weighted accuracy ($F_1$), especially when training data are scarce?

\textit{Importance}: Precision-weighted accuracy ($F_1$) or, in short, precision is/are important in contexts where false positives can cause unnecessary stress, costs, or medical interventions, emphasizing precise predictions.

\item \textbf{RQ-2}: How do LLMs compare with classical ML baselines on recall-weighted accuracy ($F_3$), especially when training data are scarce?

\textit{Importance}: Recall-weighted accuracy ($F_3$) is critical in disease diagnosis scenarios where missing positive cases (false negatives) can lead to serious health consequences.

\item \textbf{RQ-3}: Within the LLM family, which prompt-design factors (communication style, reasoning mode, shot count, external knowledge) drive performance? (\S \ref{subsec:com_style})

\textit{Importance}: Understanding how different ICL-design choices influence LLM performance can guide effective and efficient deployment of these models in clinical settings.

\item \textbf{RQ-4}: How do the same models score on group-level fairness? We quantify the demographic-parity gap, TPR gap, FPR gap, and equalised-odds distance (EOD) between male (M) and female (F) patients.

\textit{Importance}: Group-level fairness is vital for equitable healthcare. However, LLM performance in the clinical domain, especially regarding fairness, remains largely unknown.

\item \textbf{RQ-5}: How do runtime characteristics—namely training cost, per‑prediction latency, and update frequency—moderate the practical utility of fairness‑oriented LLM prompts versus classical Ml pipelines in time‑sensitive healthcare settings?
\end{itemize}

\begin{tealsection}{Experiments and Results}

We present the results of our experiments on two datasets: the Heart Disease dataset (\S\ref{subsec:health}) and the Diabetes II dataset (\S\ref{subsec:Diabetes}). For each dataset, we provide detailed answers and discussion for the research questions outlined in \S \ref{subsec:RQs}. The results are organized along the following main dimensions: (1) a comparison between ML models and LLMs in the investigated clinical downstream task, in terms of both overall accuracy ($F_1$) and risk-sensitive accuracy (measured via $F_3$); and (2) an analysis of which LLM configurations yield better performance across different domains, considering factors such as the number of examples (shots), domain knowledge integration, communication type, and use of Chain-of-Thought (CoT) prompting, and finally, (3) a deeper analysis of other risk factors, in particular the unfairness of different categories of model studied.

\subsection{Experiment I: Heart Disease Dataset}
\label{subsec:health}

We begin by discussing the results for the Heart Disease dataset. First, we examine model performance in terms of F1 score (\S\ref{subsec:health_rq1}), which is the harmonic mean of precision and recall. As discussed earlier, however, F1 alone is not sufficient to capture the cost of false predictions (for example, predicting that someone does not have a disease when they actually do). Therefore, we also provide a detailed analysis using the F3 metric (\S\ref{subsec:health_rq2}), which places greater emphasis on recall while still considering precision. These two subsections focus on systematically comparing the performance of ML models and LLMs. We repeat each analysis: once on the full dataset and once on a random sample of 50\% of the data, to assess the impact of sample size. For completeness, we present the absolute performance values in Table \ref{tab:metrics_both_datasets}, along with the corresponding performance graphs in Figures \ref{fig:health-f1} and \ref{fig:health-f3}.

\subsubsection{RQ-1: ML vs. LLMs on Precision balanced $\mathbf{F_1}$.}
\label{subsec:health_rq1}

Figure \ref{fig:health-f1} illustrates the precision-balanced F1 performance for the full Heart-disease cohort and its 50\% down-sampled subset. Notably, classical ML methods outperform LLM-based approaches under this metric. Gradient Boosting achieves the highest overall F1 score of \textbf{0.901 [0.815, 0.970]}, closely followed by Random Forest and SVM, with F1 scores of \textbf{0.869 [0.764, 0.947]} and \textbf{0.859 [0.754, 0.943]}, respectively. In comparison, the top-performing LLM variant (NL-ST, 16-shot, direct, with domain knowledge) achieves an F1 of \textbf{0.860 [0.767, 0.937]}, approximately 4.8\%  behind the leading ML method.

Interestingly, the performance gap between ML and LLM methods narrows significantly with increased shot count and natural-language prompting styles. Specifically, LLM variants with 8–16 balanced shots combined with a natural-language single-turn template approach ML performance, while numeric multi-turn prompts consistently underperform by 6–10 percentage points.

Moreover, data scarcity disproportionately impacts classical ML methods. When the data is down-sampled by half, the $F_1$ performance of Random Forest decreases to \textbf{0.830 [0.720, 0.921]}, while the best-performing LLM drops to \textbf{0.768 [0.657, 0.864]}. Although ML methods continue to maintain higher $F_1$ scores (e.g., RF: 0.830 vs. LLM: 0.768), more LLM-based methods begin to appear in the top five (NL-ST, 8- and 16-shot, direct, with domain knowledge), and at least three other models outperform Gradient Boosting (which performed best in the full-data scenario).

\begin{figure}[!h]
    \centering
    \begin{subfigure}[t]{0.70\textwidth}
        \centering
        \includegraphics[
            width=\linewidth,
            trim=40 0 0 0, 
            clip
        ]{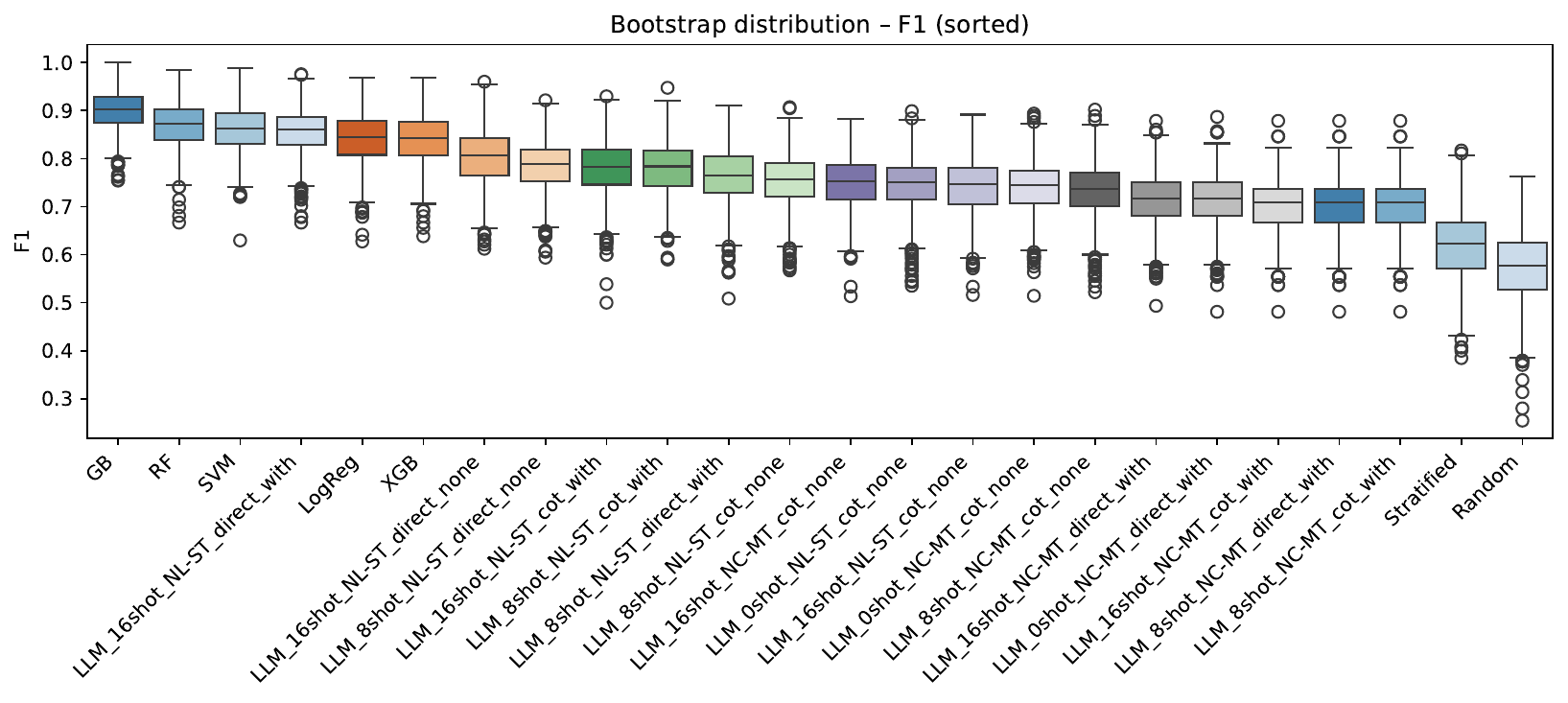}
        \caption{Full Dataset: $F_1$}
    \end{subfigure}

    \begin{subfigure}[t]{0.70\textwidth}
        \centering
        \includegraphics[
            width=\linewidth,
            trim=40 0 0 0, 
            clip
        ]{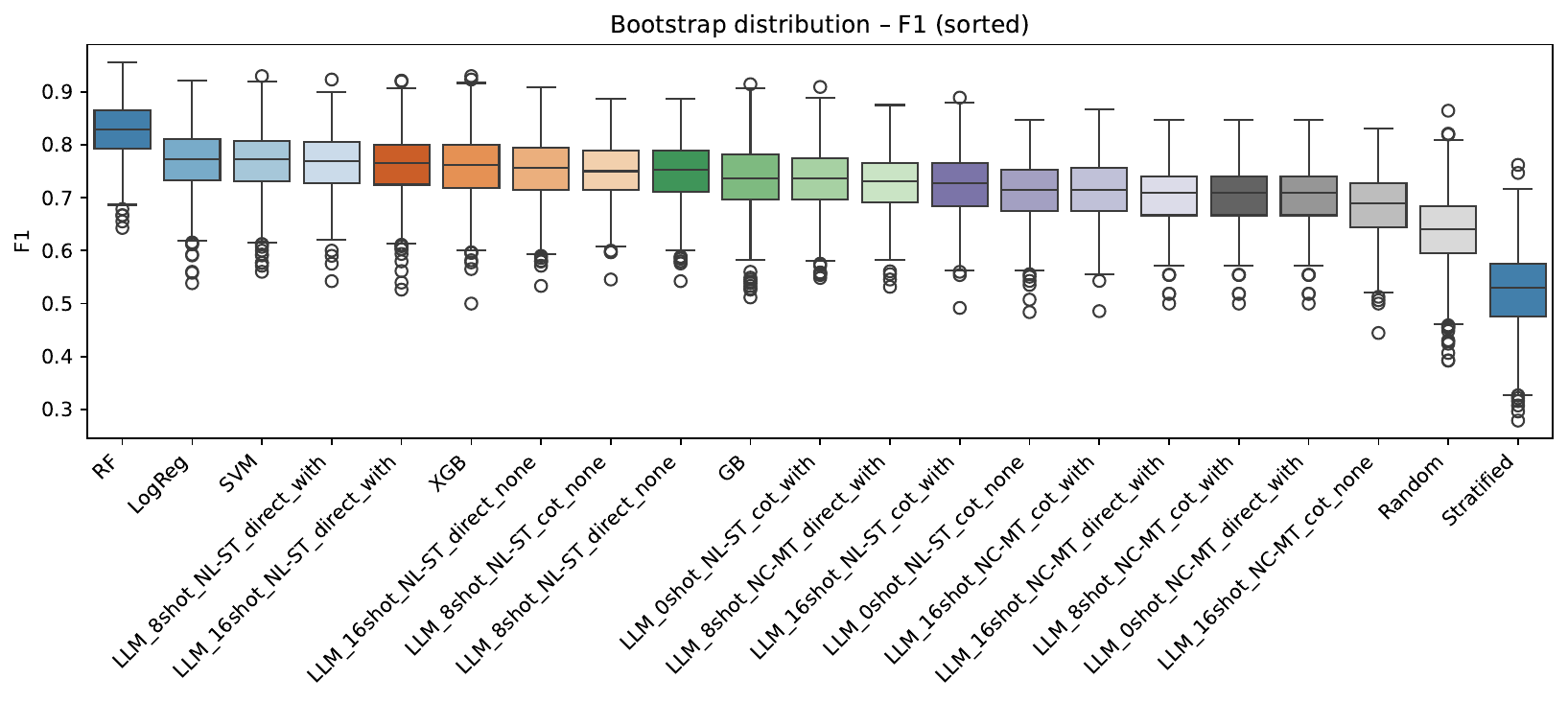}
        \caption{Sampled Dataset: $F_1$}
    \end{subfigure}

    \caption{Performance of top models on the recall-weighted metric \textbf{$F_1$} for the \textbf{Heart} dataset, under both full and sampled data regimes.}
    \label{fig:health-f1}
\end{figure}

\subsubsection{RQ-2: ML vs. LLMs on Recall-Weighted $\mathbf{F_3}$.}
\label{subsec:health_rq2}

Figure \ref{fig:health-f3} presents the comparative performance of ML and LLM models on the recall-weighted $F_3$ metric, which places a stronger emphasis on recall—critical in clinical contexts where missed cases are highly consequential. In contrast to precision-balanced metrics, LLM-based approaches emerge as the leading performers under this recall-prioritized measure.

The best overall performance is delivered by the 8-shot NL-ST prompt (direct reasoning), achieving an $F_3$ of \textbf{0.923 [0.833–0.981]}, surpassing Gradient Boosting’s best score of \textbf{0.859}. Remarkably, the top-5 and even top-10 positions are all occupied by LLM variants, demonstrating a clear superiority of LLMs when recall is weighted heavily.

This superior performance can be attributed primarily to the LLMs’ improved recall despite a modest trade-off in precision. The capacity of LLM-based methods to capture more true-positive cases substantially enhances their value in clinical scenarios, where the implications of missing a critical diagnosis far outweigh the costs associated with false positives.

When the dataset is curtailed to 50\,\%, LLMs retain their superiority.  The same LLM configuration still records an $\mathrm{F}_{3}$\textbf{=0.877\,[0.777,0.955]}, whereas the top ML model under this constraint, Random Forest, has the highest performance of \textbf{0.821 = [0.686,0.928]}, showing that   Even the \emph{best} ML contender in the sampled setting (Random Forest at $0.821$) lags the LLM by 0.056. Statistical validation through the Mann–Whitney U test reveals 

Clinically, these results might suggest implementing a hybrid diagnostic pipeline, where \textit{ high-precision ML methods serve as an initial filter}, \textit{followed by high-recall LLM models for cases demanding} rigorous identification to effectively minimize false-negative outcomes.

\begin{figure}[!h]
    \centering
    \begin{subfigure}[t]{0.70\textwidth}
        \centering
        \includegraphics[
            width=\linewidth,
            trim=10 0 0 0, 
            clip
        ]{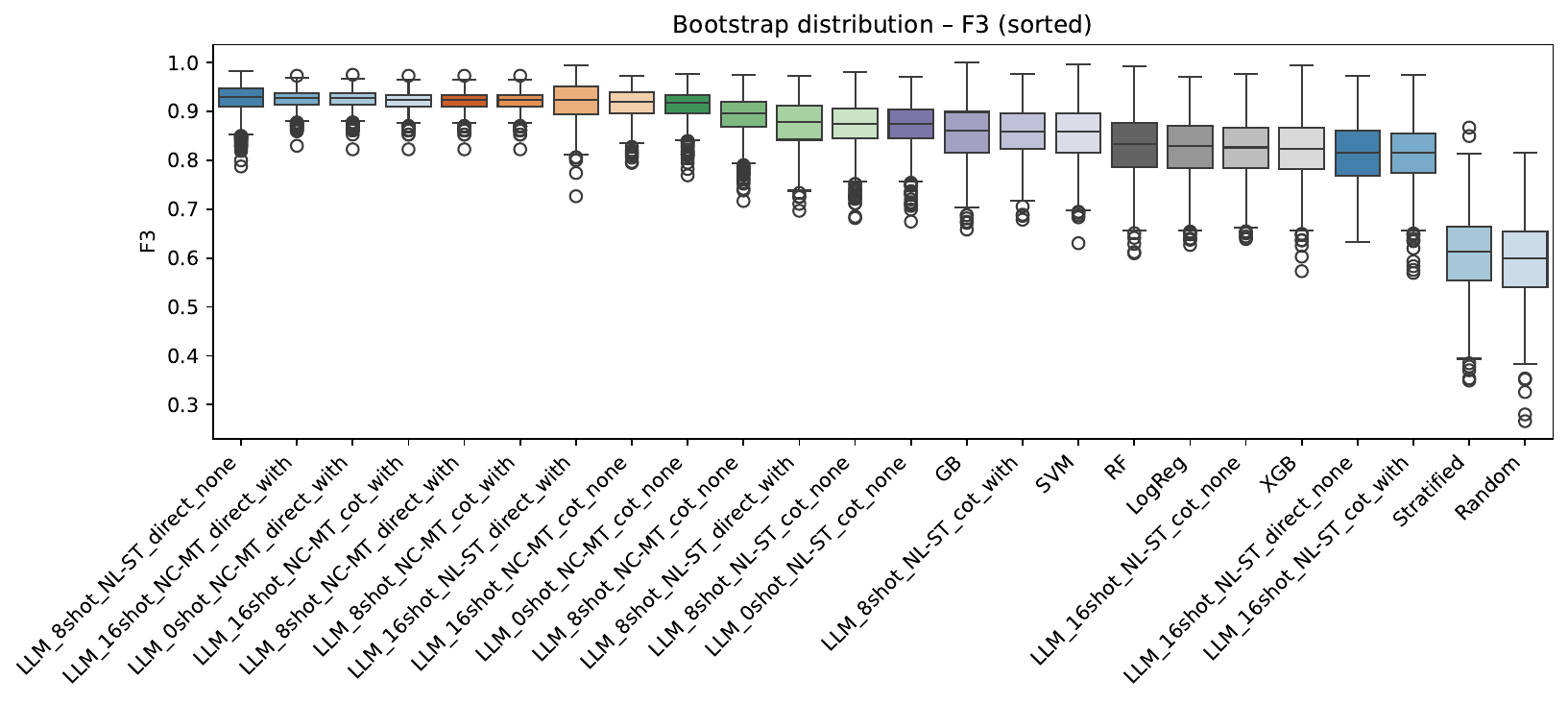}
        \caption{Full Dataset: $F_3$}
    \end{subfigure}

    \begin{subfigure}[t]{0.70\textwidth}
        \centering
        \includegraphics[
            width=\linewidth,
            trim=10 0 0 0, 
            clip
        ]{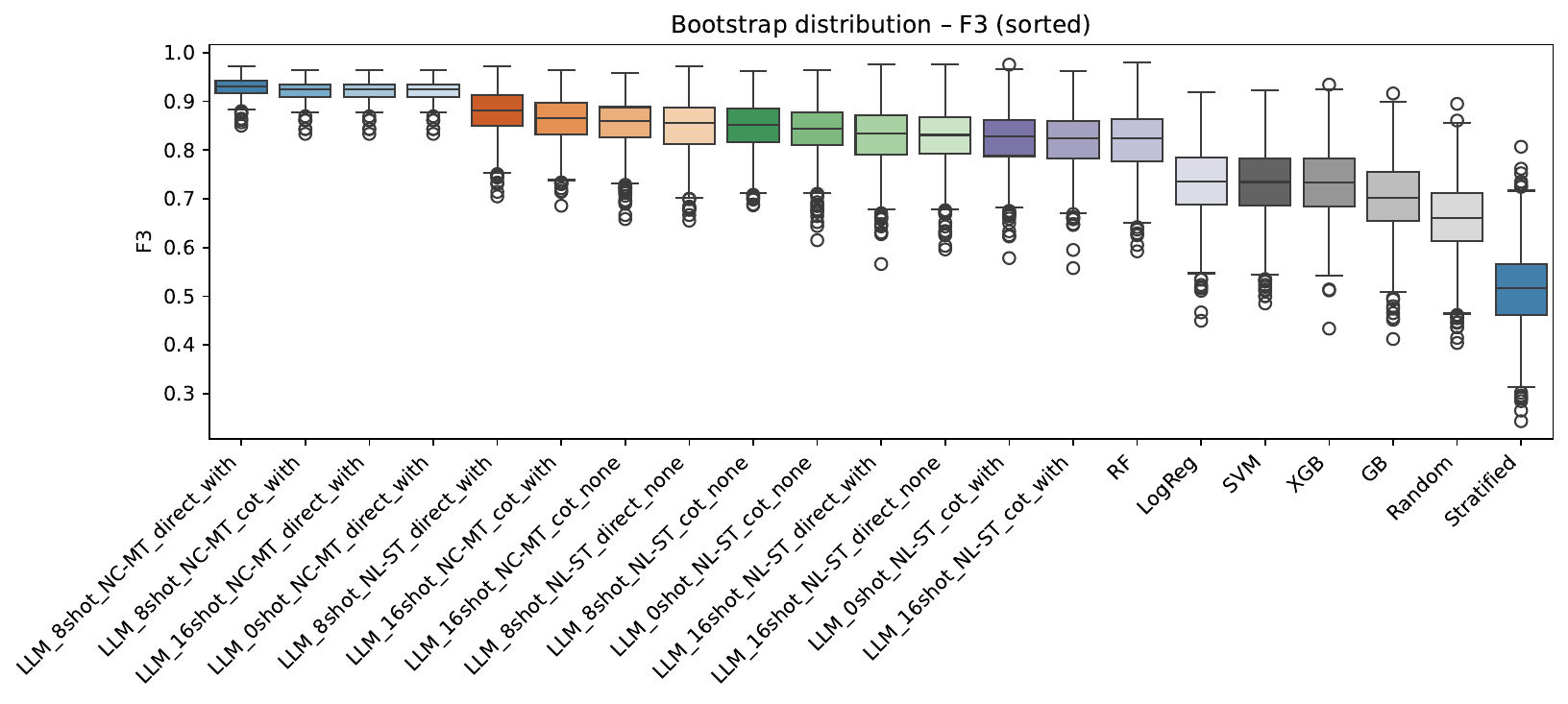}
        \caption{Sampled Dataset: $F_3$}
    \end{subfigure}

    \caption{Performance of top models on the recall-weighted metric \textbf{$F_3$} for the \textbf{Heart} dataset, under both full and sampled data regimes.}
    \label{fig:health-f3}
\end{figure}

\begin{table}[ht]
\centering
\caption{Performance metrics (Recall, Precision, F$_1$, F$_3$) for two datasets. The best value per column (within each dataset) is \textbf{bold}.}
\scriptsize

\textbf{Dataset 1: Previous Results} \\[0.5em]
\begin{tabular}{lcccc|cccc}
\toprule
& \multicolumn{4}{c|}{\textbf{Full data}} & \multicolumn{4}{c}{\textbf{Sample data}}\\
\cmidrule(r){2-5}\cmidrule(l){6-9}
\textbf{Model} & Rec. & Prec. & $F_{1}$ (CI) & $F_{3}$ (CI) & Rec. & Prec. & $F_{1}$ (CI) & $F_{3}$ (CI)\\
\midrule
GB              & 0.849 & \textbf{0.964} & \textbf{0.901} [0.815, 0.970] & 0.859 [0.735, 0.969] & 0.698 & 0.796 & 0.741 [0.600, 0.853] & 0.706 [0.543, 0.848]\\
RF              & 0.818 & 0.930 & 0.869 [0.764, 0.947] & 0.828 [0.691, 0.941] & 0.819 & \textbf{0.845} & 0.830 [0.720, 0.921] & 0.821 [0.686, 0.928]\\
SVM             & 0.848 & 0.874 & 0.859 [0.754, 0.943] & 0.850 [0.719, 0.954] & 0.728 & 0.829 & 0.772 [0.638, 0.879] & 0.736 [0.580, 0.867]\\
XGB             & 0.819 & 0.870 & 0.842 [0.733, 0.928] & 0.823 [0.687, 0.936] & 0.727 & 0.802 & 0.760 [0.640, 0.870] & 0.733 [0.588, 0.867]\\
LogReg          & 0.818 & 0.870 & 0.841 [0.733, 0.928] & 0.822 [0.687, 0.936] & 0.728 & 0.829 & 0.772 [0.636, 0.878] & 0.736 [0.575, 0.871]\\
\midrule
LLM (max)       & \textbf{0.940} & 0.795 & 0.860 [0.767, 0.937] & \textbf{0.923} [0.833, 0.981] & \textbf{0.910} & 0.667 & \textbf{0.768} [0.657, 0.864] & \textbf{0.877} [0.777, 0.955]\\
LLM (avg)       & 0.838 & 0.608 & 0.688 [0.602, 0.762] & 0.800 [0.757, 0.835] & 0.786 & 0.599 & 0.663 [0.571, 0.737] & 0.755 [0.699, 0.803]\\
\midrule
Stratified      & 0.603 & 0.641 & 0.618 [0.471, 0.747] & 0.606 [0.441, 0.765] & 0.515 & 0.548 & 0.527 [0.375, 0.667] & 0.517 [0.350, 0.677]\\
Random          & 0.606 & 0.554 & 0.575 [0.421, 0.704] & 0.599 [0.431, 0.748] & 0.667 & 0.611 & 0.634 [0.491, 0.758] & 0.659 [0.499, 0.811]\\
\bottomrule
\end{tabular}

\vspace{1.4em}

\textbf{Dataset 2: Diabatter} \\[0.5em]
\begin{tabular}{lcccc|cccc}
\toprule
& \multicolumn{4}{c|}{\textbf{Full data}} & \multicolumn{4}{c}{\textbf{Sample data}}\\
\cmidrule(r){2-5}\cmidrule(l){6-9}
\textbf{Model} & Rec. & Prec. & $F_{1}$ (CI) & $F_{3}$ (CI) & Rec. & Prec. & $F_{1}$ (CI) & $F_{3}$ (CI)\\
\midrule
RF              & 0.809 & 0.740 & \textbf{0.769} [0.611, 0.894] & 0.800 [0.628, 0.938] & 0.765 & 0.592 & 0.662 [0.488, 0.800] & 0.740 [0.555, 0.884]\\
GB              & 0.621 & \textbf{0.814} & 0.698 [0.514, 0.850] & 0.634 [0.429, 0.825] & 0.626 & 0.567 & 0.589 [0.400, 0.743] & 0.617 [0.405, 0.802]\\
SVM             & 0.618 & 0.813 & 0.697 [0.516, 0.851] & 0.632 [0.426, 0.829] & 0.620 & 0.651 & 0.629 [0.432, 0.783] & 0.621 [0.412, 0.805]\\
XGB             & 0.619 & 0.811 & 0.696 [0.514, 0.850] & 0.632 [0.420, 0.828] & 0.624 & 0.620 & 0.616 [0.432, 0.769] & 0.622 [0.417, 0.805]\\
LogReg          & 0.618 & 0.813 & 0.697 [0.516, 0.851] & 0.632 [0.426, 0.829] & 0.620 & \textbf{0.684} & 0.645 [0.450, 0.800] & 0.624 [0.417, 0.810]\\
\midrule
LLM (max)       & \textbf{1.000} & 0.527 & 0.686 [0.538, 0.806] & \textbf{0.914} [0.854, 0.954] & \textbf{0.813} & 0.607 & \textbf{0.690} [0.524, 0.825] & \textbf{0.784} [0.616, 0.913]\\
LLM (avg)       & 0.812 & 0.430 & 0.553 [0.432, 0.657] & 0.737 [0.661, 0.799] & 0.773 & 0.422 & 0.529 [0.412, 0.628] & 0.701 [0.635, 0.751]\\
\midrule
Stratified      & 0.239 & 0.251 & 0.241 [0.059, 0.410] & 0.239 [0.060, 0.425] & 0.286 & 0.300 & 0.288 [0.105, 0.462] & 0.286 [0.105, 0.476]\\
Random          & 0.718 & 0.419 & 0.525 [0.360, 0.667] & 0.667 [0.484, 0.830] & 0.666 & 0.387 & 0.485 [0.320, 0.639] & 0.618 [0.423, 0.789]\\
\bottomrule
\end{tabular}
\label{tab:metrics_both_datasets}
\end{table}

\subsection{Experiment II: Diabetes Dataset}
\label{subsec:Diabetes}

\subsubsection{RQ-1: ML vs.\ LLMs on Precision-Balanced $\mathbf{F_1}$ (Diabetes).}
\label{subsec:diabetes_rq1}

Figure~\ref{fig:diabette-f1}~compares the precision-balanced $\mathrm{F}_1$ performance for the full Diabetes dataset and its 50\% down-sampled subset. Classical ML models dominate under the full-data scenario. Specifically, Random Forest achieves the highest $\mathrm{F}_1$ score of \textbf{0.769\,[0.611,0.894]}, followed by Gradient Boosting and SVM, each with approximately $0.698$. The strongest LLM variant (NL-ST, 8-shot, direct) achieves a lower $\mathrm{F}_1$ of \textbf{0.686\,[0.538,0.806]}, roughly 8.3 percentage points behind the best ML model, placing outside the top-5 positions, which are all occupied by ML models.

Interestingly, the scenario changes when the data is down-sampled by half. The top-performing LLM (again NL-ST, 8-shot, direct) slightly increases to \textbf{0.690 \,[0.524,0.825]}, whereas the leading ML method (Random Forest) decreases notably to \textbf{0.662 \,[0.488,0.800]}. This corresponds to a relative performance loss of 13.9\% for ML (from $0.769$ to $0.662$), while the best LLM sees a slight relative improvement (0.6\%) under reduced data. Although ML methods maintain the majority in the top-5 ranks (four ML vs.\ one LLM), the reduced performance gap demonstrates the superior robustness and sample efficiency of LLM prompts under data-limited conditions.

\begin{figure}[!h]
    \centering

    \begin{subfigure}[t]{0.75\textwidth}
        \centering
    \hspace{+12mm}
        \includegraphics[
            width=\linewidth,
            trim=2 0 0 0,
            clip
        ]{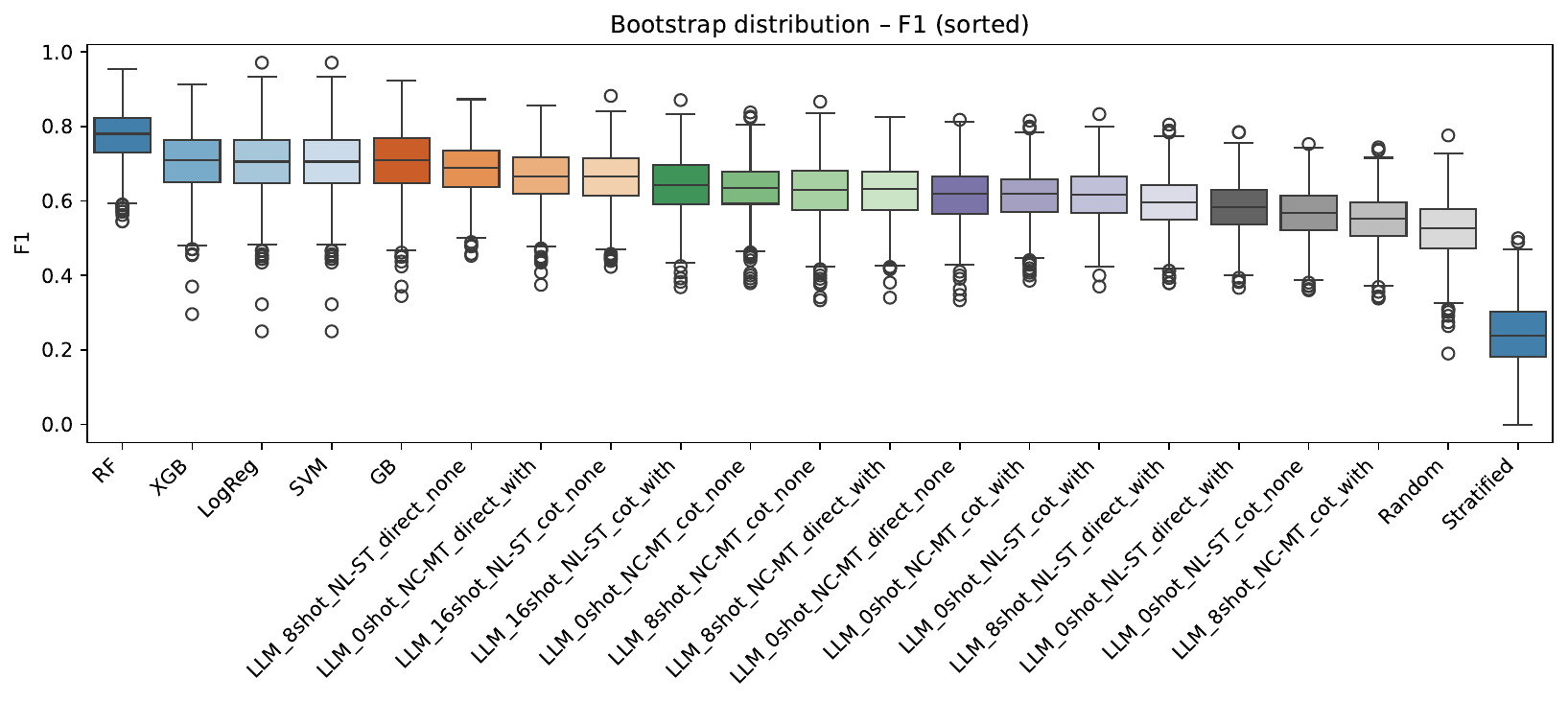}
        \caption{Full Dataset: $F_1$}
    \end{subfigure}

    \begin{subfigure}[t]{0.810\textwidth}
        \centering
    \hspace{-10mm}
        \includegraphics[
            width=\linewidth,
            trim=2 0 0 0,
            clip
        ]{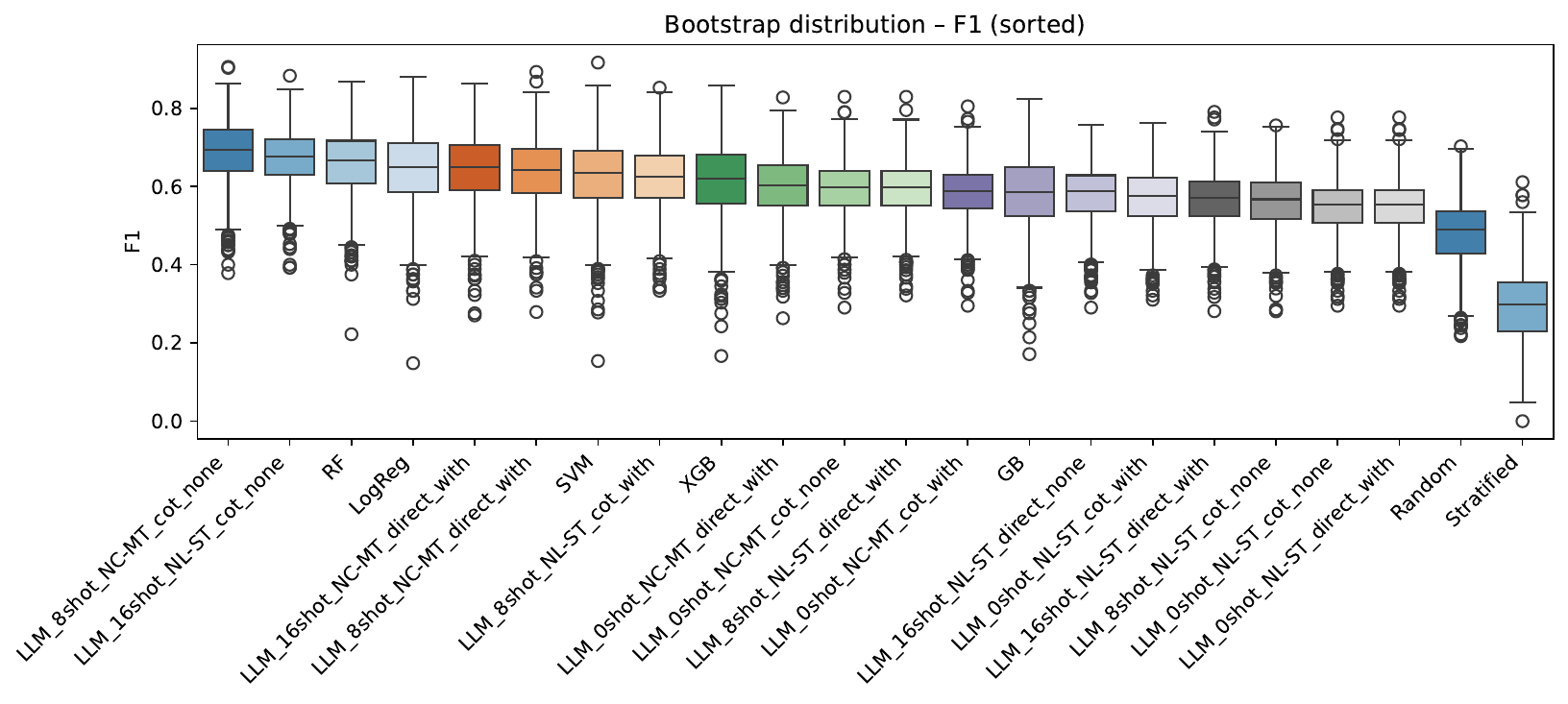}
        \caption{Sampled Dataset: $F_1$}
    \end{subfigure}

    \caption{Performance of top models on the precision balanced metric \textbf{$F_1$} for the \textbf{Diabetes} dataset, under both full and sampled data regimes.}
    \label{fig:diabette-f1}
\end{figure}

\begin{figure}[!h]
    \centering

    \begin{subfigure}[t]{0.81\textwidth}
        \centering
    \hspace{+1mm}
        \includegraphics[
            width=\linewidth,
            trim=2 0 0 0,
            clip
        ]{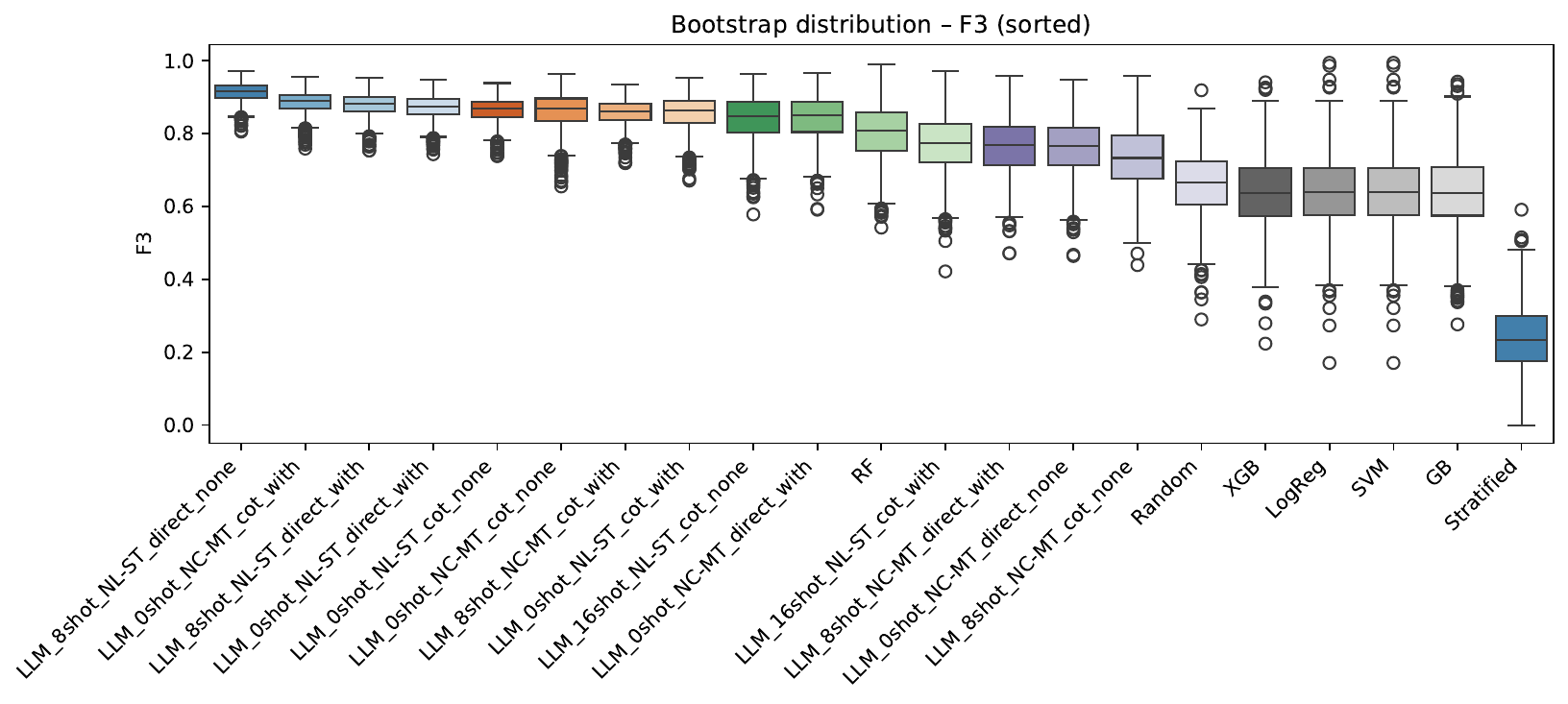}
        \caption{Full Dataset: $F_3$}
    \end{subfigure}

    \begin{subfigure}[t]{0.810\textwidth}
        \centering
    \hspace{-2mm}
        \includegraphics[
            width=\linewidth,
            trim=2 0 0 0,
            clip
        ]{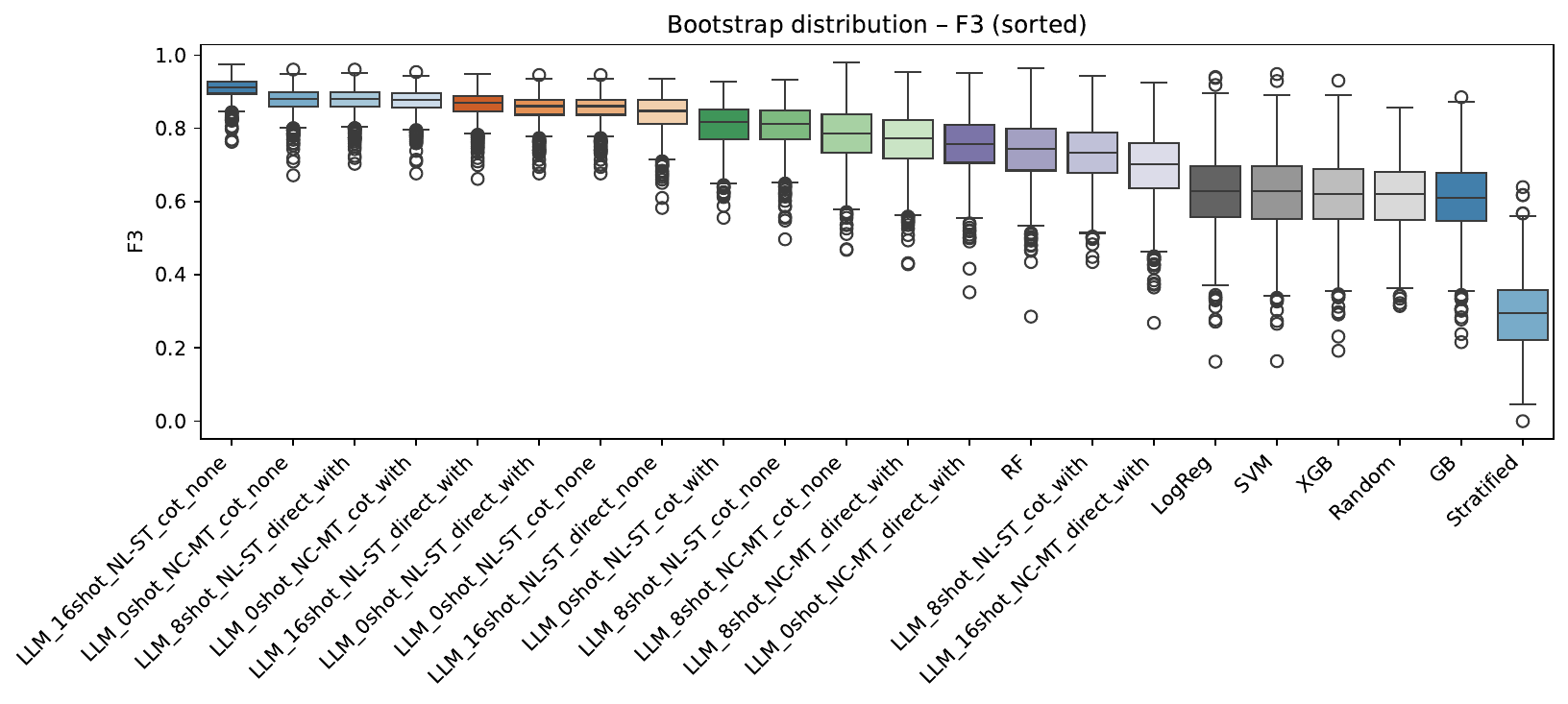}
        \caption{Sampled Dataset: $F_3$}
    \end{subfigure}

    \caption{Performance of top models on the recall-weighted metric \textbf{$F_3$} for the \textbf{Diabetes} dataset, under both full and sampled data regimes.}
    \label{fig:diabette-f1}
\end{figure}

\subsubsection{RQ-2: ML vs. LLMs on Recall-Weighted $\mathbf{F_3}$ (Diabetes)}
\label{subsec:diabetes_rq2}

Figure~\ref{fig:diabette-f1} shows comparative performance on the recall-weighted $\mathrm{F}_3$ metric, which prioritizes recall heavily, making it particularly relevant for clinical scenarios. Under this metric, LLM-based approaches significantly outperform classical ML models in both full and down-sampled settings. Specifically, the best overall performer on the full dataset is the LLM variant (NL-ST, 8-shot, direct) achieving an impressive $\mathrm{F}_3$ of \textbf{0.914\,[0.854,0.954]}, substantially higher than the best ML competitor (Random Forest) at \textbf{0.800\,[0.628,0.938]}. Notably, all top-5 positions are occupied by LLM-based variants, underscoring their dominance under recall-centric evaluation.

When the dataset is down-sampled to 50\%, the best LLM variant remains ahead with an $\mathrm{F}_3$ of \textbf{0.784\,[0.616,0.913]}, while Random Forest decreases to \textbf{0.740\,[0.555,0.884]}. In relative terms, the LLM experiences a performance drop of approximately 14.2\%, slightly larger than the ML model’s drop of 7.5\%. Nevertheless, the absolute advantage persists strongly in favor of the LLM, which continues to exclusively occupy the top-5 rankings. 

Clinically, this indicates an effective two-stage workflow for diabetes diagnosis: an initial stage employing a classical ML model to quickly exclude clear negatives, followed by a second stage leveraging high-recall LLM prompts to rigorously identify positive cases and minimize clinically costly false negatives.

\subsection{RQ3. Impact of ICL Design Factors}
\label{sec:combined-correlation}

We systematically investigated how different aspects of in-context Learning (ICL) design factors powered by LLMs influence performance across two medical classification tasks: \textbf{Heart-Disease} and \textbf{Diabetes-II}. For each dataset, we defined performance in two ways:

\begin{enumerate}
    \item \textbf{Continuous rank $\uparrow$}: Lower rank indicates better performance.
    \item \textbf{Binary success (Top-10) ($\downarrow$)}: Achieving rank within the top-10 models.
\end{enumerate}

\noindent Four key ICL design factors were studied:

\begin{itemize}
    \item \textbf{Communication Style (\texttt{comm})}: NL-ST (natural-language single-task = 1), NC-MT (non-conversational multi-task = 0).
    \item \textbf{Number of Few-shot Examples (\texttt{shots})}: Numeric values (0, 8, 16).
    \item \textbf{Reasoning Type (\texttt{reasoning})}: Chain-of-Thought (CoT = 1), Direct reasoning (0).
    \item \textbf{Knowledge Integration (\texttt{knowledge})}: With knowledge context (1), without knowledge context (0).
\end{itemize}

\subsubsection{Methodology}
\label{subsec:method}

Two correlation measures quantified these relationships:

\begin{itemize}
    \item \textbf{Spearman's rank correlation coefficient ($\rho$)} for continuous ranks:
    \[
    \rho = 1 - \frac{6 \sum d_i^2}{n(n^2-1)}
    \]
    where $d_i$ is the difference in ranks, and $n$ is the number of observations.

    \item \textbf{Point-biserial correlation ($r_{pb}$)} for binary success (Top-10 inclusion):
    \[
    r_{pb} = \frac{\bar{x}_1 - \bar{x}_0}{s_x} \sqrt{\frac{n_1 n_0}{n^2}}
    \]
    where $\bar{x}_1$, $\bar{x}_0$ are means of the continuous variable for the success/no-success groups, $s_x$ is the standard deviation, and $n_1, n_0$ are group sizes.
\end{itemize}

\noindent We present results from individual factor correlation analyses, examining how each factor—communication, shots, reasoning, and additional knowledge integration—relates to the outcomes (ranking and binary success), both in isolation (\S\ref{subsubsec:ind_fact}) and through their pairwise interactions (\S\ref{subsubsec:pair_fact}).

\subsubsection{Individual Factor Correlation Results.}
\label{subsubsec:ind_fact}
In Table~\ref{tab:indiv-corr-new}, we present the individual factor correlations for both datasets, calculated according to the formulas provided in Subsection~\ref{subsec:method}.

\begin{table}[h]
\caption{Individual factor correlations with performance measures obtained using $F_1$ figures on Full datasets.}
\centering\scriptsize
\begin{tabular}{l
                p{1.2cm}
                p{1.2cm}
                p{1.2cm}
                p{1.2cm}
                p{3.9cm}}
\toprule
\textbf{Aspect} & 
\multicolumn{2}{c}{\textbf{Heart}} & 
\multicolumn{2}{c}{\textbf{Diabetes}} & 
\multirow{2}{*}{\textbf{Observations}} \\
 & $\rho$ (Rank~$\downarrow$) & $r$ (Top-10~$\uparrow$) & $\rho$ (Rank~$\downarrow$) & $r$ (Top-10~$\uparrow$) & \\
\midrule
NL-ST style        & \textbf{--0.82} & \textbf{+0.61} & --0.04 & +0.22 & Strong impact for Heart; mild for Diabetes\\
Shots (\#examples) & --0.28 & +0.32 & \textbf{--0.31} & \textbf{+0.34} & Consistently moderate benefit across both\\
{CoT reasoning} & +0.20 & --0.25 & +0.12 & +0.10 & Mildly harmful for Heart; neutral/slightly beneficial for Diabetes\\
Knowledge context  & +0.24 & +0.09 & +0.29 & --0.22 & Neutral to mildly negative\\
\bottomrule
\end{tabular}
\label{tab:indiv-corr-new}
\end{table}

We summarize the main insights from the individual factor analysis according to the following dimensions. Please note that all these insights are based on the $F_1$ evaluation measure.

\begin{itemize}
    \item The NL-ST communication style strongly benefits Heart-Disease but has minimal effect for Diabetes-II.
    \item Increasing the number of few-shot examples consistently enhances performance on both tasks.
    \item CoT reasoning tends to slightly harm Heart performance, particularly Top-10 probability, while having a neutral or small positive effect on Diabetes.
    \item Knowledge integration alone generally has negligible benefits or even slight detrimental effects.
\end{itemize}

\subsubsection{Pairwise Factor Interaction Results}
\label{subsubsec:pair_fact}

We now move to the pairwise correlation analysis. Table~\ref{tab:pairwise-corr-new} presents the pairwise interactions between each of the above factor dimensions.

\begin{table}[h]
\centering\scriptsize
\begin{tabular}{l
                p{1.2cm}
                p{1.2cm}
                p{1.2cm}
                p{1.2cm}
                p{4cm}}
\toprule
\textbf{Interaction} & 
\multicolumn{2}{c}{\textbf{Heart}} & 
\multicolumn{2}{c}{\textbf{Diabetes}} & 
\multirow{2}{*}{\textbf{Observations}} \\
 & $\rho$ (Rank~$\downarrow$) & $r$ (Top-10~$\uparrow$) & $\rho$ (Rank~$\downarrow$) & $r$ (Top-10~$\uparrow$) & \\
\midrule
\textbf{style $\times$ shots} & \textbf{--0.80} & \textbf{+0.70} & \textbf{--0.46} & \textbf{+0.57} & Strongest beneficial interaction for both datasets\\
style $\times$ cot     & --0.26 & +0.15 & --0.10 & +0.32 & Mildly beneficial when paired with NL-ST\\
style $\times$ knowledge  & --0.57 & +0.55 & +0.20 & --0.06 & Beneficial only for Heart\\
shots $\times$ cot     & +0.12 & --0.08 & --0.17 & +0.36 & Mixed effects across datasets\\
shots $\times$ knowledge  & +0.06 & +0.23 & +0.20 & +0.02 & Minimal or neutral impact\\
cot $\times$ knowledge & +0.14 & +0.25 & +0.24 & --0.06 & Neutral to slightly negative effects\\
\bottomrule
\end{tabular}
\caption{Pairwise interaction correlations with performance measures (updated reasoning encoding).}
\label{tab:pairwise-corr-new}
\end{table}

The main insights from pairwise interactions are as follows:

\begin{itemize}
    \item \textbf{style $\times$ shots.} The interaction of NL-ST style and a high number of shots remains the most robust combination across both datasets.
    \item \textbf{style $\times$ cot} CoT reasoning merged with NL-ST provides mild positive effects, especially noticeable for Diabetes-II.
\item Knowledge shows a clear positive effect when combined with style (\textbf{style $\times$ knowledge}) on the Heart dataset, but not for the Diabetes dataset. It is the COT with shot or COT with style that has a moderate positive effect on Diabetes II.

\end{itemize}

\section*{Commonalities and Differences Across Tasks}

Across both heart-disease and diabetes classification, performance rises most reliably when the prompt includes \textbf{more balanced few-shot examples}, and the \textbf{natural-language, single-turn (NL-ST) format} continues to be the easiest conduit for the model to absorb them. Yet the way each design lever interacts with the data differs: for heart disease, NL-ST alone yields a marked boost, whereas the same change is neutral for diabetes; conversely, \textbf{chain-of-thought (CoT) and explicit feature-importance cues are not universally ``bad''}---they are simply \emph{conditional}. CoT nudges recall upward when paired with NL-ST on diabetes, and domain knowledge substantially increases precision when merged with NL-ST on heart disease, underscoring that these ingredients matter most when they are \emph{aligned with the right communication style and task} rather than when introduced in isolation.

Our work indicates several practical guidelines for ICL and prompt design:
\begin{enumerate}
    \item \textbf{Start with shots, then style.} Increase the number of balanced examples first; once you reach 8--16, switch to NL-ST so the model can process them as a coherent block.
    \item \textbf{Layer knowledge and CoT judiciously.} Add feature-importance or CoT cues only after stabilizing the prompt; each can enhance results, but \emph{only when matched to the task and style}: knowledge aids NL-ST on heart disease, while CoT aids NL-ST on diabetes.
    \item \textbf{Optimize, do not penalize.} Treat domain knowledge and CoT as \emph{tunable} assets---they can drive improvements when tailored to the precision or recall requirements of the downstream task;
    \item \textbf{Iterate empirically.} As the value of each ICL design is context-dependent, we recommend running ablation studies---varying shots, style, CoT, and knowledge---to develop a robust prompt that integrates both structured knowledge and reasoning.
\end{enumerate}

\subsection{RQ4. Fairness Analysis of the Top-Performing LLM Model}

In Table~\ref{tab:group_fairness_gaps}, we provide a summary of group-fairness gap metrics across all LLM models (approximately 20 setups), as well as the average and individual values for the ML models.

\begin{table}[ht]
\caption{Group–fairness gap metrics: demographic-parity difference ($\mathrm{DP}_{\mathrm{diff}}$), true-positive-rate difference ($\mathrm{TPR}_{\mathrm{diff}}$), false-positive-rate difference ($\mathrm{FPR}_{\mathrm{diff}}$), and equalised-odds distance ($\mathrm{EOD}$). Shaded rows summarise all LLMs and classical ML models.}

\footnotesize
\centering
\rowcolors{3}{}{gray!8}
\begin{tabular}{lcccccccc}
\toprule
Model & Shots & Comm\_Style & Reasoning & Domain\_Knowledge & $\mathrm{DP}_{\mathrm{diff}}$ & $\mathrm{TPR}_{\mathrm{diff}}$ & $\mathrm{FPR}_{\mathrm{diff}}$ & $\mathrm{EOD}$ \\
\midrule
GB & - & - & - & - & 0.5370 & 0.8750 & 0.0455 & 0.4602 \\
RF & - & - & - & - & 0.5370 & 0.8438 & 0.0909 & 0.4673 \\
XGB & - & - & - & - & 0.5741 & 0.8438 & 0.1818 & 0.5128 \\
LogReg & - & - & - & - & 0.5741 & 0.8438 & 0.1818 & 0.5128 \\
SVM & - & - & - & - & 0.5926 & 0.8750 & 0.1818 & 0.5284 \\
Stratified & - & - & - & - & -0.1667 & 0.6250 & -0.4818 & 0.5534 \\
Random & - & - & - & - & 0.1111 & -0.4062 & 0.2364 & 0.3213 \\
\midrule
\rowcolor{gray!20}
\textbf{LLM mean} &  &  &  &  & 0.0463 & -0.0391 & -0.0561 & 0.1271 \\
\rowcolor{gray!20}
\textbf{LLM median} &  &  &  &  & 0.0093 & -0.0625 & -0.0500 & 0.1171 \\
\rowcolor{cyan!15}
\textbf{ML mean} &  &  &  &  & 0.3942 & 0.6429 & 0.0623 & 0.4795 \\
\rowcolor{cyan!15}
\textbf{ML median} &  &  &  &  & 0.5370 & 0.8438 & 0.1818 & 0.5128 \\
\bottomrule
\end{tabular}
\label{tab:group_fairness_gaps}
\end{table}

\paragraph{Interpretation of Fairness Metrics.}
The table reports four standard group–fairness gap metrics:
\begin{enumerate}
    \item \textbf{Demographic Parity difference} ($\mathrm{DP}_{\mathrm{diff}}$): The difference in positive prediction rates across demographic groups. Zero indicates perfect fairness.
    \item \textbf{True-Positive Rate difference} ($\mathrm{TPR}_{\mathrm{diff}}$): Disparity in correct positive predictions between groups. Zero is ideal.
    \item \textbf{False-Positive Rate difference} ($\mathrm{FPR}_{\mathrm{diff}}$): Difference in false positives. Zero means no group disparity.
    \item \textbf{Equalized Odds Distance} ($\mathrm{EOD}$): Aggregates TPR and FPR disparities; lower is better.
\end{enumerate}

Focusing on the \emph{average} and \emph{median} values in Table~\ref{tab:group_fairness_gaps}, our results suggest a consistent fairness advantage for Large Language Models (LLMs) over the suite of classical machine‑learning (ML) baselines.

First, the \textbf{demographic‑parity difference} for LLMs is close to neutral on both statistics (\(\overline{\mathrm{DP}_{\mathrm{diff}}}=0.0463\); \(\widetilde{\mathrm{DP}_{\mathrm{diff}}}=0.0093\)), whereas ML models cluster around substantially higher—and uniformly positive—gaps (\(\overline{\mathrm{DP}_{\mathrm{diff}}}=0.3942\); \(\widetilde{\mathrm{DP}_{\mathrm{diff}}}=0.5370\)).  In practical terms, LLM predictions deviate only marginally from parity, while classical models exhibit a systematic tilt favouring the unprotected group.  

A similar pattern emerges for the \textbf{true‑positive‑rate difference} and \textbf{false‑positive‑rate difference}.  On average, LLMs record a slight negative \(\mathrm{TPR}_{\mathrm{diff}}\) (\(-0.0391\)) and \(\mathrm{FPR}_{\mathrm{diff}}\) (\(-0.0561\)), signalling a minor bias in favour of the protected group; by contrast, ML baselines overshoot in the opposite direction, with mean gaps of \(0.6429\) and \(0.0623\), respectively.  Although the sign flips, the magnitude gap remains large, reinforcing that traditional models amplify disparities that LLMs nearly cancel out.

Finally, the composite \textbf{equalised‑odds distance} further underscores the divide: the mean EOD for LLMs (\(0.1271\)) is roughly one‑quarter of the ML mean (\(0.4795\)), and the median follows suit (\(0.1171\) vs.\ \(0.5128\)).  Recalling the fairness semantics introduced in Section~\ref{sec:fairness-metrics}, these lower EOD values indicate that LLMs move markedly closer to simultaneous parity in both error rates.

Taken together, our work indicates that, \emph{in the aggregate}, knowledge‑integrated LLMs deliver substantially fairer outcomes than classical ML pipelines—even before any dedicated bias‑mitigation step is applied.  This finding aligns with the broader observation that models which “reason” through chain‑of‑thought and incorporate domain cues can balance predictive accuracy with equity, whereas accuracy‑optimised but knowledge‑agnostic algorithms often leave sizeable fairness gaps unaddressed.

\subsection{Runtime Trade‑Offs: LLMs vs.\ Classical Machine‑Learning Pipelines}

A detailed examination of Table~\ref{tab:timing_comparison} demonstrates that Large Language Models (LLMs) exhibit substantially higher \emph{per-prediction} latency, exceeding the sub-millisecond inference times of classical machine-learning models by approximately two to three orders of magnitude. The longest LLM execution time (626 seconds) is associated with prompts that integrate chain-of-thought reasoning alongside comprehensive domain knowledge, while the shortest execution (approximately 24 seconds) is observed with more succinct, straightforward prompts. Since LLMs function exclusively at inference time, their overall computational cost is determined entirely by the duration of a single forward pass. Consequently, even the median LLM latency—approximately two minutes—may present a significant obstacle for real-time or time-sensitive clinical applications.

In contrast, classical machine-learning pipelines display an inverse computational profile. Their inference phase is effectively instantaneous, rendering them highly suitable for operational environments that demand rapid response times. However, the training phase can become prohibitively time-consuming when employing sophisticated algorithms, such as support vector machines or ensemble methods; for example, training the SVM model on the full dataset requires nearly ten minutes. In contexts where models are trained once and subsequently deployed at scale, the initial training overhead is quickly amortised. Conversely, in scenarios necessitating frequent model retraining, the cumulative training time may approach or even exceed the total inference time required by LLMs.

It is also important to note that LLMs, by design, eliminate the need for any separate training phase at deployment: once the foundational model is instantiated, all learning and inference occur simultaneously via zero-shot or few-shot prompting. As a result, the reported inference latency for LLMs fully encapsulates their deployment cost, whereas classical machine-learning systems invariably require the additional consideration of both training and inference durations.

\begin{table}[!h]
\caption{Wall–clock time comparison between Large Language Models (LLMs) and classical machine‑learning (ML) pipelines on the \textit{heart‑disease} task.  LLM values report the end‑to‑end latency of a single prompt execution (no separate training phase).  ML values are split into model–training and model–inference stages.  All numbers in seconds; ‘‘min–max’’ indicates the fastest and slowest configurations inside each block.}
\footnotesize
\centering
\begin{tabular}{@{}lccccc@{}}
\toprule
\textbf{Family} & \textbf{Phase} & \textbf{Dataset} & \textbf{Mean} & \textbf{Median} & \textbf{Min--Max} \\
\midrule
LLM             & Prompt exec.\  & Full    & 172.0 & 197.4 & 23.6--626.4 \\
LLM             & Prompt exec.\  & Sample  & 152.4 & 127.6 & 24.8--365.4 \\
\midrule
Classical ML    & Training       & Full    & 68.4  & 3.83  & 0.0004--577.7 \\
Classical ML    & Training       & Sample  & 35.1  & 0.87  & 0.0003--282.5 \\
Classical ML    & Inference      & Full    & 0.0033 & 0.0022 & 0.00014--0.0147 \\
Classical ML    & Inference      & Sample  & 0.0031 & 0.0015 & 0.00011--0.0160 \\
\bottomrule
\end{tabular}
\label{tab:timing_comparison}
\end{table}

\section{Conclusion and Future Work}
This work presents a comprehensive evaluation of a knowledge-guided in-context learning (ICL) pipeline, leveraging Large Language Models (LLMs) for structured clinical prediction tasks. By systematically integrating domain knowledge, balanced few-shot exemplars, and tailored prompt engineering—including both natural-language single-turn (NL-ST) and numeric conversational multi-turn (NC-MT) styles—our approach enables LLMs to closely match, and in high-recall settings surpass, the performance of traditional machine learning (ML) models such as Gradient Boosting and Random Forest. Notably, the pipeline delivers state-of-the-art recall for critical diagnoses while also achieving significantly improved gender fairness, reducing demographic parity and equalized odds gaps by an order of magnitude compared to ML baselines. The analysis further clarifies key prompt design levers and quantifies the trade-off between the LLMs’ strong diagnostic sensitivity and their substantially higher inference latency. For a concise overview of all research questions addressed, we refer readers to the summary tables provided in the following above.

Looking forward, future research will focus on distilling the most effective, high-recall LLM prompts into compact student models to enable real-time deployment and reduce latency. We also plan to expand the knowledge-injection paradigm to handle multi-modal electronic health record data and to extend fairness assessments to intersectional dimensions such as age and ethnicity. These directions aim to preserve the diagnostic and equity advantages observed while making the system practical for broader clinical adoption.

Last but not least, one limitation, arguably, is that our current evaluation is fully automated and does not include human expert assessment. As outlined in the revised Limitations and Future Directions, we plan to conduct larger-scale human evaluation studies in subsequent work. We hope these additions provide the qualitative depth the reviewer requested and lay a strong foundation for future human-centered research.

\begin{tcolorbox}[colback=blue!6!white, colframe=blue!65!black, title={Summary of Answers to Research Questions: RQ1--RQ3}]
\begin{itemize}
    \item \textbf{RQ-1 (Precision-weighted accuracy, F$_1$):} 
    Classical machine learning (ML) models generally outperform large language models (LLMs) in balanced precision and recall. However, the gap narrows significantly when LLMs utilize balanced few-shot contexts and natural language narrative prompts (NL-ST). 

    \item \textbf{RQ-2 (Recall-weighted accuracy, F$_3$):} 
    When prioritizing recall (minimizing missed diagnoses), knowledge-guided LLMs decisively outperform classical ML models. For example, on heart disease, an 8-shot NL-ST LLM achieves F$_3$ = 0.923, exceeding Gradient Boosting’s 0.859. Similarly, for diabetes, the LLM scores 0.914 versus Random Forest’s 0.800.

    \item \textbf{RQ-3 (Impact of In-Context Learning (ICL) design):} 
    Four prompt-engineering factors critically affect performance:
    \begin{enumerate}
        \item \emph{Communication format:} NL-ST maximizes recall; 
        \item \emph{Number of shots:} Increasing from 0 to 8 to 16 examples steadily improves both F$_1$ and F$_3$.
        \item \emph{Reasoning method:} Chain-of-Thought (CoT) with NL-ST yields highest recall;
        \item \emph{Domain knowledge inclusion:} Adding dominant and moderate feature buckets improves metrics; minor features add noise.
    \end{enumerate}
    These findings are supported by correlation analyses across 70 prompt variants.
\end{itemize}
\end{tcolorbox}

\vspace{1em}

\begin{tcolorbox}[colback=gray!7!white, colframe=gray!65!black, title={Summary of Answers to Research Questions: RQ4--RQ5}]
\begin{itemize}
    \item \textbf{RQ-4 (Fairness across gender):} 
    Knowledge-integrated LLMs demonstrate substantially improved fairness metrics compared to classical ML models. \\
    \textit{Evidence:} Median demographic parity gap reduces from 0.537 (ML) to 0.009 (LLM), and Equalized Odds distance reduces from 0.513 to 0.117, effectively mitigating male-advantage bias in false negatives.

    \item \textbf{RQ-5 (Runtime and deployment trade-offs):} 
    LLMs provide zero-training flexibility but incur significantly higher inference latency. In contrast, classical ML models require one-time training but offer sub-millisecond inference suitable for real-time applications. \\
    \textit{Example:} Heart disease inference with LLM prompts averages 197 seconds (range 24--626 s), whereas classical ML inference averages 0.002 seconds, with training times up to 577 seconds (SVM).
\end{itemize}
\end{tcolorbox}

\end{tealsection}


\bibliographystyle{ACM-Reference-Format}
\bibliography{refs}


\begin{thebibliography}{46}


\ifx \showCODEN    \undefined \def \showCODEN     #1{\unskip}     \fi
\ifx \showDOI      \undefined \def \showDOI       #1{#1}\fi
\ifx \showISBNx    \undefined \def \showISBNx     #1{\unskip}     \fi
\ifx \showISBNxiii \undefined \def \showISBNxiii  #1{\unskip}     \fi
\ifx \showISSN     \undefined \def \showISSN      #1{\unskip}     \fi
\ifx \showLCCN     \undefined \def \showLCCN      #1{\unskip}     \fi
\ifx \shownote     \undefined \def \shownote      #1{#1}          \fi
\ifx \showarticletitle \undefined \def \showarticletitle #1{#1}   \fi
\ifx \showURL      \undefined \def \showURL       {\relax}        \fi
\providecommand\bibfield[2]{#2}
\providecommand\bibinfo[2]{#2}
\providecommand\natexlab[1]{#1}
\providecommand\showeprint[2][]{arXiv:#2}

\bibitem[Amugongo et~al\mbox{.}(2025)]%
        {amugongo2025retrieval}
\bibfield{author}{\bibinfo{person}{Lameck~Mbangula Amugongo}, \bibinfo{person}{Pietro Mascheroni}, \bibinfo{person}{Steven Brooks}, \bibinfo{person}{Stefan Doering}, {and} \bibinfo{person}{Jan Seidel}.} \bibinfo{year}{2025}\natexlab{}.
\newblock \showarticletitle{Retrieval augmented generation for large language models in healthcare: A systematic review}.
\newblock \bibinfo{journal}{\emph{PLOS Digital Health}} \bibinfo{volume}{4}, \bibinfo{number}{6} (\bibinfo{year}{2025}), \bibinfo{pages}{e0000877}.
\newblock


\bibitem[Chen and Esmaeilzadeh(2024)]%
        {chen2024generative}
\bibfield{author}{\bibinfo{person}{Yan Chen} {and} \bibinfo{person}{Pouyan Esmaeilzadeh}.} \bibinfo{year}{2024}\natexlab{}.
\newblock \showarticletitle{Generative AI in medical practice: in-depth exploration of privacy and security challenges}.
\newblock \bibinfo{journal}{\emph{Journal of Medical Internet Research}}  \bibinfo{volume}{26} (\bibinfo{year}{2024}), \bibinfo{pages}{e53008}.
\newblock


\bibitem[Choi et~al\mbox{.}(2017)]%
        {DBLP:conf/mlhc/ChoiBMDSS17}
\bibfield{author}{\bibinfo{person}{Edward Choi}, \bibinfo{person}{Siddharth Biswal}, \bibinfo{person}{Bradley~A. Malin}, \bibinfo{person}{Jon Duke}, \bibinfo{person}{Walter~F. Stewart}, {and} \bibinfo{person}{Jimeng Sun}.} \bibinfo{year}{2017}\natexlab{}.
\newblock \showarticletitle{Generating Multi-label Discrete Patient Records using Generative Adversarial Networks}. In \bibinfo{booktitle}{\emph{Proceedings of the Machine Learning for Health Care Conference, {MLHC} 2017, Boston, Massachusetts, USA, 18-19 August 2017}} \emph{(\bibinfo{series}{Proceedings of Machine Learning Research}, Vol.~\bibinfo{volume}{68})}, \bibfield{editor}{\bibinfo{person}{Finale Doshi{-}Velez}, \bibinfo{person}{Jim Fackler}, \bibinfo{person}{David~C. Kale}, \bibinfo{person}{Rajesh Ranganath}, \bibinfo{person}{Byron~C. Wallace}, {and} \bibinfo{person}{Jenna Wiens}} (Eds.). \bibinfo{publisher}{{PMLR}}, \bibinfo{pages}{286--305}.
\newblock
\urldef\tempurl%
\url{http://proceedings.mlr.press/v68/choi17a.html}
\showURL{%
\tempurl}


\bibitem[Deldjoo(2023)]%
        {deldjoo2023fairnesschatgpt}
\bibfield{author}{\bibinfo{person}{Yashar Deldjoo}.} \bibinfo{year}{2023}\natexlab{}.
\newblock \showarticletitle{Fairness of ChatGPT and the Role Of Explainable-Guided Prompts}.
\newblock \bibinfo{journal}{\emph{arXiv preprint arXiv:2307.11761}} (\bibinfo{year}{2023}).
\newblock
\urldef\tempurl%
\url{https://arxiv.org/abs/2307.11761}
\showURL{%
\tempurl}


\bibitem[Deldjoo(2024)]%
        {deldjoo2024fairevalllm}
\bibfield{author}{\bibinfo{person}{Yashar Deldjoo}.} \bibinfo{year}{2024}\natexlab{}.
\newblock \showarticletitle{FairEvalLLM. A Comprehensive Framework for Benchmarking Fairness in Large Language Model Recommender Systems}.
\newblock \bibinfo{journal}{\emph{arXiv preprint arXiv:2405.02219}} (\bibinfo{year}{2024}).
\newblock


\bibitem[Deldjoo et~al\mbox{.}(2024a)]%
        {deldjoo2024review}
\bibfield{author}{\bibinfo{person}{Yashar Deldjoo}, \bibinfo{person}{Zhankui He}, \bibinfo{person}{Julian McAuley}, \bibinfo{person}{Anton Korikov}, \bibinfo{person}{Scott Sanner}, \bibinfo{person}{Arnau Ramisa}, \bibinfo{person}{Ren{\'e} Vidal}, \bibinfo{person}{Maheswaran Sathiamoorthy}, \bibinfo{person}{Atoosa Kasirzadeh}, {and} \bibinfo{person}{Silvia Milano}.} \bibinfo{year}{2024}\natexlab{a}.
\newblock \showarticletitle{A Review of Modern Recommender Systems Using Generative Models (Gen-RecSys)}.
\newblock \bibinfo{journal}{\emph{arXiv preprint arXiv:2404.00579}} (\bibinfo{year}{2024}).
\newblock


\bibitem[Deldjoo et~al\mbox{.}(2024b)]%
        {deldjoo2024recommendation}
\bibfield{author}{\bibinfo{person}{Yashar Deldjoo}, \bibinfo{person}{Zhankui He}, \bibinfo{person}{Julian McAuley}, \bibinfo{person}{Anton Korikov}, \bibinfo{person}{Scott Sanner}, \bibinfo{person}{Arnau Ramisa}, \bibinfo{person}{Rene Vidal}, \bibinfo{person}{Maheswaran Sathiamoorthy}, \bibinfo{person}{Atoosa Kasrizadeh}, \bibinfo{person}{Silvia Milano}, {et~al\mbox{.}}} \bibinfo{year}{2024}\natexlab{b}.
\newblock \showarticletitle{Recommendation with Generative Models}.
\newblock \bibinfo{journal}{\emph{arXiv preprint arXiv:2409.15173}} (\bibinfo{year}{2024}).
\newblock


\bibitem[Galitsky(2024)]%
        {galitsky2024llm}
\bibfield{author}{\bibinfo{person}{Boris~A Galitsky}.} \bibinfo{year}{2024}\natexlab{}.
\newblock \showarticletitle{LLM-Based Personalized Recommendations in Health}.
\newblock  (\bibinfo{year}{2024}).
\newblock


\bibitem[Ge et~al\mbox{.}(2023)]%
        {DBLP:journals/jbi/GeGDAS23}
\bibfield{author}{\bibinfo{person}{Yao Ge}, \bibinfo{person}{Yuting Guo}, \bibinfo{person}{Sudeshna Das}, \bibinfo{person}{Mohammed~Ali Al{-}Garadi}, {and} \bibinfo{person}{Abeed Sarker}.} \bibinfo{year}{2023}\natexlab{}.
\newblock \showarticletitle{Few-shot learning for medical text: {A} review of advances, trends, and opportunities}.
\newblock \bibinfo{journal}{\emph{J. Biomed. Informatics}}  \bibinfo{volume}{144} (\bibinfo{year}{2023}), \bibinfo{pages}{104458}.
\newblock
\urldef\tempurl%
\url{https://doi.org/10.1016/J.JBI.2023.104458}
\showDOI{\tempurl}


\bibitem[Ghosh et~al\mbox{.}(2024)]%
        {ghosh2024clipsyntel}
\bibfield{author}{\bibinfo{person}{Akash Ghosh}, \bibinfo{person}{Arkadeep Acharya}, \bibinfo{person}{Raghav Jain}, \bibinfo{person}{Sriparna Saha}, \bibinfo{person}{Aman Chadha}, {and} \bibinfo{person}{Setu Sinha}.} \bibinfo{year}{2024}\natexlab{}.
\newblock \showarticletitle{Clipsyntel: clip and llm synergy for multimodal question summarization in healthcare}. In \bibinfo{booktitle}{\emph{Proceedings of the AAAI Conference on Artificial Intelligence}}, Vol.~\bibinfo{volume}{38}. \bibinfo{pages}{22031--22039}.
\newblock


\bibitem[Goyal et~al\mbox{.}(2024)]%
        {goyal2024healai}
\bibfield{author}{\bibinfo{person}{Sagar Goyal}, \bibinfo{person}{Eti Rastogi}, \bibinfo{person}{Sree~Prasanna Rajagopal}, \bibinfo{person}{Dong Yuan}, \bibinfo{person}{Fen Zhao}, \bibinfo{person}{Jai Chintagunta}, \bibinfo{person}{Gautam Naik}, {and} \bibinfo{person}{Jeff Ward}.} \bibinfo{year}{2024}\natexlab{}.
\newblock \showarticletitle{HealAI: A Healthcare LLM for Effective Medical Documentation}. In \bibinfo{booktitle}{\emph{Proceedings of the 17th ACM International Conference on Web Search and Data Mining}}. \bibinfo{pages}{1167--1168}.
\newblock


\bibitem[Gramopadhye et~al\mbox{.}(2024)]%
        {gramopadhye2024few}
\bibfield{author}{\bibinfo{person}{Ojas Gramopadhye}, \bibinfo{person}{Saeel~Sandeep Nachane}, \bibinfo{person}{Prateek Chanda}, \bibinfo{person}{Ganesh Ramakrishnan}, \bibinfo{person}{Kshitij~Sharad Jadhav}, \bibinfo{person}{Yatin Nandwani}, \bibinfo{person}{Dinesh Raghu}, {and} \bibinfo{person}{Sachindra Joshi}.} \bibinfo{year}{2024}\natexlab{}.
\newblock \showarticletitle{Few shot chain-of-thought driven reasoning to prompt LLMs for open ended medical question answering}.
\newblock \bibinfo{journal}{\emph{arXiv preprint arXiv:2403.04890}} (\bibinfo{year}{2024}).
\newblock


\bibitem[Griot et~al\mbox{.}(2025)]%
        {griot2025large}
\bibfield{author}{\bibinfo{person}{Maxime Griot}, \bibinfo{person}{Coralie Hemptinne}, \bibinfo{person}{Jean Vanderdonckt}, {and} \bibinfo{person}{Demet Yuksel}.} \bibinfo{year}{2025}\natexlab{}.
\newblock \showarticletitle{Large language models lack essential metacognition for reliable medical reasoning}.
\newblock \bibinfo{journal}{\emph{Nature communications}} \bibinfo{volume}{16}, \bibinfo{number}{1} (\bibinfo{year}{2025}), \bibinfo{pages}{642}.
\newblock


\bibitem[Harshvardhan et~al\mbox{.}(2020)]%
        {harshvardhan2020comprehensive}
\bibfield{author}{\bibinfo{person}{GM Harshvardhan}, \bibinfo{person}{Mahendra~Kumar Gourisaria}, \bibinfo{person}{Manjusha Pandey}, {and} \bibinfo{person}{Siddharth~Swarup Rautaray}.} \bibinfo{year}{2020}\natexlab{}.
\newblock \showarticletitle{A comprehensive survey and analysis of generative models in machine learning}.
\newblock \bibinfo{journal}{\emph{Computer Science Review}}  \bibinfo{volume}{38} (\bibinfo{year}{2020}), \bibinfo{pages}{100285}.
\newblock


\bibitem[He et~al\mbox{.}(2023)]%
        {he2023large}
\bibfield{author}{\bibinfo{person}{Zhankui He}, \bibinfo{person}{Zhouhang Xie}, \bibinfo{person}{Rahul Jha}, \bibinfo{person}{Harald Steck}, \bibinfo{person}{Dawen Liang}, \bibinfo{person}{Yesu Feng}, \bibinfo{person}{Bodhisattwa~Prasad Majumder}, \bibinfo{person}{Nathan Kallus}, {and} \bibinfo{person}{Julian McAuley}.} \bibinfo{year}{2023}\natexlab{}.
\newblock \showarticletitle{Large language models as zero-shot conversational recommenders}. In \bibinfo{booktitle}{\emph{Proceedings of the 32nd ACM international conference on information and knowledge management}}. \bibinfo{pages}{720--730}.
\newblock


\bibitem[Henriksson et~al\mbox{.}(2023)]%
        {DBLP:journals/artmed/HenrikssonPHN23}
\bibfield{author}{\bibinfo{person}{Aron Henriksson}, \bibinfo{person}{Yash Pawar}, \bibinfo{person}{Pontus Hedberg}, {and} \bibinfo{person}{Pontus Naucl{\'{e}}r}.} \bibinfo{year}{2023}\natexlab{}.
\newblock \showarticletitle{Multimodal fine-tuning of clinical language models for predicting {COVID-19} outcomes}.
\newblock \bibinfo{journal}{\emph{Artif. Intell. Medicine}}  \bibinfo{volume}{146} (\bibinfo{year}{2023}), \bibinfo{pages}{102695}.
\newblock
\urldef\tempurl%
\url{https://doi.org/10.1016/J.ARTMED.2023.102695}
\showDOI{\tempurl}


\bibitem[Kojima et~al\mbox{.}(2022)]%
        {kojima2022large}
\bibfield{author}{\bibinfo{person}{Takeshi Kojima}, \bibinfo{person}{Shixiang~Shane Gu}, \bibinfo{person}{Machel Reid}, \bibinfo{person}{Yutaka Matsuo}, {and} \bibinfo{person}{Yusuke Iwasawa}.} \bibinfo{year}{2022}\natexlab{}.
\newblock \showarticletitle{Large language models are zero-shot reasoners}.
\newblock \bibinfo{journal}{\emph{Advances in neural information processing systems}}  \bibinfo{volume}{35} (\bibinfo{year}{2022}), \bibinfo{pages}{22199--22213}.
\newblock


\bibitem[Kusa et~al\mbox{.}(2023)]%
        {kusa2023dr}
\bibfield{author}{\bibinfo{person}{Wojciech Kusa}, \bibinfo{person}{Edoardo Mosca}, {and} \bibinfo{person}{Aldo Lipani}.} \bibinfo{year}{2023}\natexlab{}.
\newblock \showarticletitle{“Dr LLM, what do I have?”: The Impact of User Beliefs and Prompt Formulation on Health Diagnoses}. In \bibinfo{booktitle}{\emph{Proceedings of the Third Workshop on NLP for Medical Conversations}}. \bibinfo{pages}{13--19}.
\newblock


\bibitem[Li et~al\mbox{.}(2023)]%
        {li2023prompt}
\bibfield{author}{\bibinfo{person}{Lei Li}, \bibinfo{person}{Yongfeng Zhang}, {and} \bibinfo{person}{Li Chen}.} \bibinfo{year}{2023}\natexlab{}.
\newblock \showarticletitle{Prompt distillation for efficient llm-based recommendation}. In \bibinfo{booktitle}{\emph{Proceedings of the 32nd ACM International Conference on Information and Knowledge Management}}. \bibinfo{pages}{1348--1357}.
\newblock


\bibitem[Li et~al\mbox{.}(2021)]%
        {li2021survey}
\bibfield{author}{\bibinfo{person}{Zewen Li}, \bibinfo{person}{Fan Liu}, \bibinfo{person}{Wenjie Yang}, \bibinfo{person}{Shouheng Peng}, {and} \bibinfo{person}{Jun Zhou}.} \bibinfo{year}{2021}\natexlab{}.
\newblock \showarticletitle{A survey of convolutional neural networks: analysis, applications, and prospects}.
\newblock \bibinfo{journal}{\emph{IEEE transactions on neural networks and learning systems}} \bibinfo{volume}{33}, \bibinfo{number}{12} (\bibinfo{year}{2021}), \bibinfo{pages}{6999--7019}.
\newblock


\bibitem[Moor et~al\mbox{.}(2023)]%
        {DBLP:conf/ml4h/MoorHWYDLZRR23}
\bibfield{author}{\bibinfo{person}{Michael Moor}, \bibinfo{person}{Qian Huang}, \bibinfo{person}{Shirley Wu}, \bibinfo{person}{Michihiro Yasunaga}, \bibinfo{person}{Yash Dalmia}, \bibinfo{person}{Jure Leskovec}, \bibinfo{person}{Cyril Zakka}, \bibinfo{person}{Eduardo~Pontes Reis}, {and} \bibinfo{person}{Pranav Rajpurkar}.} \bibinfo{year}{2023}\natexlab{}.
\newblock \showarticletitle{Med-Flamingo: a Multimodal Medical Few-shot Learner}. In \bibinfo{booktitle}{\emph{Machine Learning for Health, ML4H@NeurIPS 2023, 10 December 2023, New Orleans, Louisiana, {USA}}} \emph{(\bibinfo{series}{Proceedings of Machine Learning Research}, Vol.~\bibinfo{volume}{225})}, \bibfield{editor}{\bibinfo{person}{Stefan Hegselmann}, \bibinfo{person}{Antonio Parziale}, \bibinfo{person}{Divya Shanmugam}, \bibinfo{person}{Shengpu Tang}, \bibinfo{person}{Mercy~Nyamewaa Asiedu}, \bibinfo{person}{Serina Chang}, \bibinfo{person}{Tom Hartvigsen}, {and} \bibinfo{person}{Harvineet Singh}} (Eds.). \bibinfo{publisher}{{PMLR}}, \bibinfo{pages}{353--367}.
\newblock
\urldef\tempurl%
\url{https://proceedings.mlr.press/v225/moor23a.html}
\showURL{%
\tempurl}


\bibitem[Nazary et~al\mbox{.}(2023)]%
        {nazary2023chatgpt}
\bibfield{author}{\bibinfo{person}{Fatemeh Nazary}, \bibinfo{person}{Yashar Deldjoo}, {and} \bibinfo{person}{Tommaso Di~Noia}.} \bibinfo{year}{2023}\natexlab{}.
\newblock \showarticletitle{ChatGPT-HealthPrompt. Harnessing the Power of XAI in Prompt-Based Healthcare Decision Support using ChatGPT}. In \bibinfo{booktitle}{\emph{European Conference on Artificial Intelligence}}. Springer, \bibinfo{pages}{382--397}.
\newblock


\bibitem[Nazi and Peng(2023)]%
        {nazi2023large}
\bibfield{author}{\bibinfo{person}{Zabir~Al Nazi} {and} \bibinfo{person}{Wei Peng}.} \bibinfo{year}{2023}\natexlab{}.
\newblock \showarticletitle{Large language models in healthcare and medical domain: A review}.
\newblock \bibinfo{journal}{\emph{arXiv preprint arXiv:2401.06775}} (\bibinfo{year}{2023}).
\newblock


\bibitem[Nori et~al\mbox{.}(2023)]%
        {nori2023capabilities}
\bibfield{author}{\bibinfo{person}{Harsha Nori}, \bibinfo{person}{Nicholas King}, \bibinfo{person}{Scott~Mayer McKinney}, \bibinfo{person}{Dean Carignan}, {and} \bibinfo{person}{Eric Horvitz}.} \bibinfo{year}{2023}\natexlab{}.
\newblock \showarticletitle{Capabilities of gpt-4 on medical challenge problems}.
\newblock \bibinfo{journal}{\emph{arXiv preprint arXiv:2303.13375}} (\bibinfo{year}{2023}).
\newblock


\bibitem[Obermeyer et~al\mbox{.}(2019)]%
        {obermeyer2019dissecting}
\bibfield{author}{\bibinfo{person}{Ziad Obermeyer}, \bibinfo{person}{Brian Powers}, \bibinfo{person}{Christine Vogeli}, {and} \bibinfo{person}{Sendhil Mullainathan}.} \bibinfo{year}{2019}\natexlab{}.
\newblock \showarticletitle{Dissecting racial bias in an algorithm used to manage the health of populations}.
\newblock \bibinfo{journal}{\emph{Science}} \bibinfo{volume}{366}, \bibinfo{number}{6464} (\bibinfo{year}{2019}), \bibinfo{pages}{447--453}.
\newblock


\bibitem[OpenAI(2022)]%
        {openai2022chatgpt}
\bibfield{author}{\bibinfo{person}{TB OpenAI}.} \bibinfo{year}{2022}\natexlab{}.
\newblock \bibinfo{title}{Chatgpt: Optimizing language models for dialogue. OpenAI}.
\newblock
\newblock


\bibitem[Peng et~al\mbox{.}(2023)]%
        {peng2023study}
\bibfield{author}{\bibinfo{person}{Cheng Peng}, \bibinfo{person}{Xi Yang}, \bibinfo{person}{Aokun Chen}, \bibinfo{person}{Kaleb~E Smith}, \bibinfo{person}{Nima PourNejatian}, \bibinfo{person}{Anthony~B Costa}, \bibinfo{person}{Cheryl Martin}, \bibinfo{person}{Mona~G Flores}, \bibinfo{person}{Ying Zhang}, \bibinfo{person}{Tanja Magoc}, {et~al\mbox{.}}} \bibinfo{year}{2023}\natexlab{}.
\newblock \showarticletitle{A study of generative large language model for medical research and healthcare}.
\newblock \bibinfo{journal}{\emph{NPJ Digital Medicine}} \bibinfo{volume}{6}, \bibinfo{number}{1} (\bibinfo{year}{2023}), \bibinfo{pages}{210}.
\newblock


\bibitem[Rajkomar et~al\mbox{.}(2018)]%
        {rajkomar2018ensuring}
\bibfield{author}{\bibinfo{person}{Alvin Rajkomar}, \bibinfo{person}{Michaela Hardt}, \bibinfo{person}{Michael~D Howell}, \bibinfo{person}{Greg Corrado}, {and} \bibinfo{person}{Marshall~H Chin}.} \bibinfo{year}{2018}\natexlab{}.
\newblock \showarticletitle{Ensuring fairness in machine learning to advance health equity}.
\newblock \bibinfo{journal}{\emph{Annals of internal medicine}} \bibinfo{volume}{169}, \bibinfo{number}{12} (\bibinfo{year}{2018}), \bibinfo{pages}{866--872}.
\newblock


\bibitem[Rezayi et~al\mbox{.}(2023)]%
        {rezayi2023exploring}
\bibfield{author}{\bibinfo{person}{Saed Rezayi}, \bibinfo{person}{Zhengliang Liu}, \bibinfo{person}{Zihao Wu}, \bibinfo{person}{Chandra Dhakal}, \bibinfo{person}{Bao Ge}, \bibinfo{person}{Haixing Dai}, \bibinfo{person}{Gengchen Mai}, \bibinfo{person}{Ninghao Liu}, \bibinfo{person}{Chen Zhen}, \bibinfo{person}{Tianming Liu}, {et~al\mbox{.}}} \bibinfo{year}{2023}\natexlab{}.
\newblock \showarticletitle{Exploring new frontiers in agricultural nlp: Investigating the potential of large language models for food applications}.
\newblock \bibinfo{journal}{\emph{arXiv preprint arXiv:2306.11892}} (\bibinfo{year}{2023}).
\newblock


\bibitem[Sarkar et~al\mbox{.}(2020)]%
        {sarkar2020machine}
\bibfield{author}{\bibinfo{person}{Suproteem~K Sarkar}, \bibinfo{person}{Subhrajit Roy}, \bibinfo{person}{Emily Alsentzer}, \bibinfo{person}{Matthew~BA McDermott}, \bibinfo{person}{Fabian Falck}, \bibinfo{person}{Ioana Bica}, \bibinfo{person}{Griffin Adams}, \bibinfo{person}{Stephen Pfohl}, {and} \bibinfo{person}{Stephanie~L Hyland}.} \bibinfo{year}{2020}\natexlab{}.
\newblock \showarticletitle{Machine learning for health (ML4H) 2020: Advancing healthcare for all}. In \bibinfo{booktitle}{\emph{Machine Learning for Health}}. PMLR, \bibinfo{pages}{1--11}.
\newblock


\bibitem[Savage et~al\mbox{.}(2024)]%
        {savage2024diagnostic}
\bibfield{author}{\bibinfo{person}{Thomas Savage}, \bibinfo{person}{Ashwin Nayak}, \bibinfo{person}{Robert Gallo}, \bibinfo{person}{Ekanath Rangan}, {and} \bibinfo{person}{Jonathan~H Chen}.} \bibinfo{year}{2024}\natexlab{}.
\newblock \showarticletitle{Diagnostic reasoning prompts reveal the potential for large language model interpretability in medicine}.
\newblock \bibinfo{journal}{\emph{NPJ Digital Medicine}} \bibinfo{volume}{7}, \bibinfo{number}{1} (\bibinfo{year}{2024}), \bibinfo{pages}{20}.
\newblock


\bibitem[Sayin et~al\mbox{.}(2024a)]%
        {sayin2024can}
\bibfield{author}{\bibinfo{person}{Burcu Sayin}, \bibinfo{person}{Pasquale Minervini}, \bibinfo{person}{Jacopo Staiano}, {and} \bibinfo{person}{Andrea Passerini}.} \bibinfo{year}{2024}\natexlab{a}.
\newblock \showarticletitle{Can LLMs Correct Physicians, Yet? Investigating Effective Interaction Methods in the Medical Domain}.
\newblock \bibinfo{journal}{\emph{arXiv preprint arXiv:2403.20288}} (\bibinfo{year}{2024}).
\newblock


\bibitem[Sayin et~al\mbox{.}(2024b)]%
        {sayin2024llmscorrectphysiciansyet}
\bibfield{author}{\bibinfo{person}{Burcu Sayin}, \bibinfo{person}{Pasquale Minervini}, \bibinfo{person}{Jacopo Staiano}, {and} \bibinfo{person}{Andrea Passerini}.} \bibinfo{year}{2024}\natexlab{b}.
\newblock \bibinfo{title}{Can LLMs Correct Physicians, Yet? Investigating Effective Interaction Methods in the Medical Domain}.
\newblock
\newblock
\showeprint[arxiv]{2403.20288}~[cs.CL]
\urldef\tempurl%
\url{https://arxiv.org/abs/2403.20288}
\showURL{%
\tempurl}


\bibitem[Shentu and Al~Moubayed(2024)]%
        {shentu2024cxr}
\bibfield{author}{\bibinfo{person}{Junjie Shentu} {and} \bibinfo{person}{Noura Al~Moubayed}.} \bibinfo{year}{2024}\natexlab{}.
\newblock \showarticletitle{CXR-IRGen: An Integrated Vision and Language Model for the Generation of Clinically Accurate Chest X-Ray Image-Report Pairs}. In \bibinfo{booktitle}{\emph{Proceedings of the IEEE/CVF Winter Conference on Applications of Computer Vision}}. \bibinfo{pages}{5212--5221}.
\newblock


\bibitem[Singhal et~al\mbox{.}(2023)]%
        {singhal2023large}
\bibfield{author}{\bibinfo{person}{Karan Singhal}, \bibinfo{person}{Shekoofeh Azizi}, \bibinfo{person}{Tao Tu}, \bibinfo{person}{S~Sara Mahdavi}, \bibinfo{person}{Jason Wei}, \bibinfo{person}{Hyung~Won Chung}, \bibinfo{person}{Nathan Scales}, \bibinfo{person}{Ajay Tanwani}, \bibinfo{person}{Heather Cole-Lewis}, \bibinfo{person}{Stephen Pfohl}, {et~al\mbox{.}}} \bibinfo{year}{2023}\natexlab{}.
\newblock \showarticletitle{Large language models encode clinical knowledge}.
\newblock \bibinfo{journal}{\emph{Nature}} \bibinfo{volume}{620}, \bibinfo{number}{7972} (\bibinfo{year}{2023}), \bibinfo{pages}{172--180}.
\newblock


\bibitem[Spola{\^{o}}r et~al\mbox{.}(2024)]%
        {DBLP:journals/mta/SpolaorLMNPTCWF24}
\bibfield{author}{\bibinfo{person}{Newton Spola{\^{o}}r}, \bibinfo{person}{Huei~Diana Lee}, \bibinfo{person}{Ana~Isabel Mendes}, \bibinfo{person}{Concei{\c{c}}{\~{a}}o~Veloso Nogueira}, \bibinfo{person}{Antonio Rafael~Sabino Parmezan}, \bibinfo{person}{Weber Shoity~Resende Takaki}, \bibinfo{person}{Cl{\'{a}}udio Saddy~Rodrigues Coy}, \bibinfo{person}{Feng~Chung Wu}, {and} \bibinfo{person}{Rui Fonseca{-}Pinto}.} \bibinfo{year}{2024}\natexlab{}.
\newblock \showarticletitle{Fine-tuning pre-trained neural networks for medical image classification in small clinical datasets}.
\newblock \bibinfo{journal}{\emph{Multim. Tools Appl.}} \bibinfo{volume}{83}, \bibinfo{number}{9} (\bibinfo{year}{2024}), \bibinfo{pages}{27305--27329}.
\newblock
\urldef\tempurl%
\url{https://doi.org/10.1007/S11042-023-16529-W}
\showDOI{\tempurl}


\bibitem[Stark et~al\mbox{.}(2019)]%
        {stark2019predicting}
\bibfield{author}{\bibinfo{person}{Gigi~F Stark}, \bibinfo{person}{Gregory~R Hart}, \bibinfo{person}{Bradley~J Nartowt}, {and} \bibinfo{person}{Jun Deng}.} \bibinfo{year}{2019}\natexlab{}.
\newblock \showarticletitle{Predicting breast cancer risk using personal health data and machine learning models}.
\newblock \bibinfo{journal}{\emph{Plos one}} \bibinfo{volume}{14}, \bibinfo{number}{12} (\bibinfo{year}{2019}), \bibinfo{pages}{e0226765}.
\newblock


\bibitem[Sweller et~al\mbox{.}(2019)]%
        {sweller2019cognitive}
\bibfield{author}{\bibinfo{person}{John Sweller}, \bibinfo{person}{Jeroen~JG Van~Merri{\"e}nboer}, {and} \bibinfo{person}{Fred Paas}.} \bibinfo{year}{2019}\natexlab{}.
\newblock \showarticletitle{Cognitive architecture and instructional design: 20 years later}.
\newblock \bibinfo{journal}{\emph{Educational psychology review}} \bibinfo{volume}{31}, \bibinfo{number}{2} (\bibinfo{year}{2019}), \bibinfo{pages}{261--292}.
\newblock


\bibitem[Team et~al\mbox{.}(2023)]%
        {team2023gemini}
\bibfield{author}{\bibinfo{person}{Gemini Team}, \bibinfo{person}{Rohan Anil}, \bibinfo{person}{Sebastian Borgeaud}, \bibinfo{person}{Yonghui Wu}, \bibinfo{person}{Jean-Baptiste Alayrac}, \bibinfo{person}{Jiahui Yu}, \bibinfo{person}{Radu Soricut}, \bibinfo{person}{Johan Schalkwyk}, \bibinfo{person}{Andrew~M Dai}, \bibinfo{person}{Anja Hauth}, {et~al\mbox{.}}} \bibinfo{year}{2023}\natexlab{}.
\newblock \showarticletitle{Gemini: a family of highly capable multimodal models}.
\newblock \bibinfo{journal}{\emph{arXiv preprint arXiv:2312.11805}} (\bibinfo{year}{2023}).
\newblock


\bibitem[Torfi and Fox(2020)]%
        {DBLP:conf/flairs/TorfiF20}
\bibfield{author}{\bibinfo{person}{Amirsina Torfi} {and} \bibinfo{person}{Edward~A. Fox}.} \bibinfo{year}{2020}\natexlab{}.
\newblock \showarticletitle{CorGAN: Correlation-Capturing Convolutional Generative Adversarial Networks for Generating Synthetic Healthcare Records}. In \bibinfo{booktitle}{\emph{Proceedings of the Thirty-Third International Florida Artificial Intelligence Research Society Conference, Originally to be held in North Miami Beach, Florida, USA, May 17-20, 2020}}, \bibfield{editor}{\bibinfo{person}{Roman Bart{\'{a}}k} {and} \bibinfo{person}{Eric Bell}} (Eds.). \bibinfo{publisher}{{AAAI} Press}, \bibinfo{pages}{335--340}.
\newblock
\urldef\tempurl%
\url{https://aaai.org/ocs/index.php/FLAIRS/FLAIRS20/paper/view/18458}
\showURL{%
\tempurl}


\bibitem[Umer and Adnan(2024)]%
        {umer2024generative}
\bibfield{author}{\bibinfo{person}{Fahad Umer} {and} \bibinfo{person}{Niha Adnan}.} \bibinfo{year}{2024}\natexlab{}.
\newblock \showarticletitle{Generative artificial intelligence: synthetic datasets in dentistry}.
\newblock \bibinfo{journal}{\emph{BDJ open}} \bibinfo{volume}{10}, \bibinfo{number}{1} (\bibinfo{year}{2024}), \bibinfo{pages}{13}.
\newblock


\bibitem[Wu et~al\mbox{.}(2023)]%
        {wu2023bloomberggpt}
\bibfield{author}{\bibinfo{person}{Shijie Wu}, \bibinfo{person}{Ozan Irsoy}, \bibinfo{person}{Steven Lu}, \bibinfo{person}{Vadim Dabravolski}, \bibinfo{person}{Mark Dredze}, \bibinfo{person}{Sebastian Gehrmann}, \bibinfo{person}{Prabhanjan Kambadur}, \bibinfo{person}{David Rosenberg}, {and} \bibinfo{person}{Gideon Mann}.} \bibinfo{year}{2023}\natexlab{}.
\newblock \showarticletitle{Bloomberggpt: A large language model for finance}.
\newblock \bibinfo{journal}{\emph{arXiv preprint arXiv:2303.17564}} (\bibinfo{year}{2023}).
\newblock


\bibitem[Xi et~al\mbox{.}(2023)]%
        {xi2023rise}
\bibfield{author}{\bibinfo{person}{Zhiheng Xi}, \bibinfo{person}{Wenxiang Chen}, \bibinfo{person}{Xin Guo}, \bibinfo{person}{Wei He}, \bibinfo{person}{Yiwen Ding}, \bibinfo{person}{Boyang Hong}, \bibinfo{person}{Ming Zhang}, \bibinfo{person}{Junzhe Wang}, \bibinfo{person}{Senjie Jin}, \bibinfo{person}{Enyu Zhou}, {et~al\mbox{.}}} \bibinfo{year}{2023}\natexlab{}.
\newblock \showarticletitle{The rise and potential of large language model based agents: A survey}.
\newblock \bibinfo{journal}{\emph{arXiv preprint arXiv:2309.07864}} (\bibinfo{year}{2023}).
\newblock


\bibitem[Yang et~al\mbox{.}(2023)]%
        {yang2023large}
\bibfield{author}{\bibinfo{person}{Rui Yang}, \bibinfo{person}{Ting~Fang Tan}, \bibinfo{person}{Wei Lu}, \bibinfo{person}{Arun~James Thirunavukarasu}, \bibinfo{person}{Daniel Shu~Wei Ting}, {and} \bibinfo{person}{Nan Liu}.} \bibinfo{year}{2023}\natexlab{}.
\newblock \showarticletitle{Large language models in health care: Development, applications, and challenges}.
\newblock \bibinfo{journal}{\emph{Health Care Science}} \bibinfo{volume}{2}, \bibinfo{number}{4} (\bibinfo{year}{2023}), \bibinfo{pages}{255--263}.
\newblock


\bibitem[Yuan et~al\mbox{.}(2024)]%
        {yuan2024automated}
\bibfield{author}{\bibinfo{person}{Han Yuan}, \bibinfo{person}{Kunyu Yu}, \bibinfo{person}{Feng Xie}, \bibinfo{person}{Mingxuan Liu}, {and} \bibinfo{person}{Shenghuan Sun}.} \bibinfo{year}{2024}\natexlab{}.
\newblock \showarticletitle{Automated machine learning with interpretation: A systematic review of methodologies and applications in healthcare}.
\newblock \bibinfo{journal}{\emph{Medicine Advances}} \bibinfo{volume}{2}, \bibinfo{number}{3} (\bibinfo{year}{2024}), \bibinfo{pages}{205--237}.
\newblock


\bibitem[Zhao et~al\mbox{.}(2023)]%
        {zhao2023survey}
\bibfield{author}{\bibinfo{person}{Wayne~Xin Zhao}, \bibinfo{person}{Kun Zhou}, \bibinfo{person}{Junyi Li}, \bibinfo{person}{Tianyi Tang}, \bibinfo{person}{Xiaolei Wang}, \bibinfo{person}{Yupeng Hou}, \bibinfo{person}{Yingqian Min}, \bibinfo{person}{Beichen Zhang}, \bibinfo{person}{Junjie Zhang}, \bibinfo{person}{Zican Dong}, {et~al\mbox{.}}} \bibinfo{year}{2023}\natexlab{}.
\newblock \showarticletitle{A survey of large language models}.
\newblock \bibinfo{journal}{\emph{arXiv preprint arXiv:2303.18223}} (\bibinfo{year}{2023}).
\newblock


\end{thebibliography}
\end{document}